\documentclass[lettersize,journal]{IEEEtran}
\usepackage{amsmath,amsfonts}
\usepackage{algorithmic}
\usepackage{algorithm}
\usepackage{array}
\usepackage[caption=false,font=normalsize,labelfont=sf,textfont=sf]{subfig}
\usepackage{textcomp}
\usepackage{stfloats}
\usepackage{url}
\usepackage{verbatim}
\usepackage{graphicx}
\usepackage{cite}
\usepackage{lineno}
\usepackage{multirow}
\usepackage{makecell}
\usepackage{soul}
\usepackage{xspace}
\usepackage{color}
\usepackage{tabularx} 
\usepackage{ragged2e} 
\usepackage{booktabs} 
\usepackage{hyperref}

\begin{document}

\title{S2R-Bench: A Sim-to-Real Evaluation Benchmark for Autonomous Driving}

\author{Li Wang$^*$, Guangqi Yang$^*$, Lei Yang$^*$, Ziying Song$^*$, Xinyu Zhang, Ying Chen, Lin Liu, Junjie Gao, Zhiwei Li, Qingshan Yang, Jun Li, Liangliang Wang, Wenhao Yu, Bin Xu, Weida Wang, Huaping Liu

\thanks{This work was supported by the National High Technology Research and Development Program of China under Grant No.2020YFC1512500, the National Natural Science Foundation of China under Grant No. 62273198, 52221005, U22B2052, the Beijing Natural Science Foundation Program under Grant No.L241017, and the National Foundation Singapore under its AI Singapore Program under Award AISG2-GC-2023-007. \emph{(Corresponding author: Xinyu Zhang (e-mail: xyzhang@tsinghua.edu.cn), *: These authors contributed equally to this work.).}}

\thanks{Li Wang, Bin Xu and Weida Wang are with School of Machanical Engineering, Beijing Institute of Technology, Beijing 100081, China.}

\thanks{Guangqi Yang is with the College of Computer Science and Technology, Jilin University, Changchun 130012, China.}

\thanks{Lei Yang is with the School of Mechanical and Aerospace Engineering, Nanyang Technological University, Singapore 639798, Singapore.}

\thanks{Ziying Song and Lin Liu are with the School of Computer and Information Technology, Beijing Key Lab of Traffic Data Analysis and Mining, Beijing Jiaotong University, Beijing 100044, China.}

\thanks{Xinyu Zhang is with the School of Vehicle and Mobility and Suzhou Automotive Research Institute, Tsinghua University, Beijing 100084, China, and also with the Department of Electrical and Computer Engineering, National University of Singapore, Singapore 119077, Singapore.}

\thanks{Ying Chen, Junjie Gao and Jun Li are with the School of Vehicle and Mobility, Tsinghua University, Beijing 100084, China.}

\thanks{Zhiwei Li is with the College of Information Science and Technology, Beijing University of Chemical Technology, Beijing 100029, China.}

\thanks{Qingshan Yang is with the Beijing Jingwei Hirain Technologies Co., Inc., Beijing 100191, China.}

\thanks{Liangliang Wang is with the School of Computer Science and Artificial Intelligence, Wuhan University of Technology, Wuhan 430070, China.} 

\thanks{Wenhao Yu is with the Beiiing Tsinghua Institute for Frontier Interdisciplinary Innovation, Beijing 100084, China.} 

\thanks{Huaping Liu is with the State Key Laboratory of Intelligent Technology and Systems, and Department of Computer Science and Technology, Tsinghua University, Beijing 100084, China.}
}



\maketitle

\begin{abstract}
Safety is a long-standing and the final pursuit in the development of autonomous driving systems, with a significant portion of safety challenge arising from perception. How to effectively evaluate the safety as well as the reliability of perception algorithms is becoming an emerging issue. Despite its critical importance, existing perception methods exhibit a limitation in their robustness, primarily due to the use of benchmarks are entierly simulated, which fail to align predicted results with actual outcomes, particularly under extreme weather conditions and sensor anomalies that are prevalent in real-world scenarios. To fill this gap, in this study, we propose a Sim-to-Real Evaluation Benchmark for Autonomous Driving (S2R-Bench). We collect diverse sensor anomaly data under various road conditions to evaluate the robustness of autonomous driving perception methods in a comprehensive and realistic manner. This is the first corruption robustness benchmark based on real-world scenarios, encompassing various road conditions, weather conditions, lighting intensities, and time periods. By comparing real-world data with simulated data, we demonstrate the reliability and practical significance of the collected data for real-world applications. We hope that this dataset will advance future research and contribute to the development of more robust perception models for autonomous driving. This dataset is released on https://github.com/adept-thu/S2R-Bench.
\end{abstract}


\section{Introduction}
The perception challenge in autonomous driving is inherently linked to the safety and reliability of the system. The corruption robustness benchmark provides a crucial foundation to develop accurate and resilient perception methods, which are essential for ensuring the safety of autonomous vehicles\cite{yu2024online,mahima2024toward,liu2024benchmarking,he2023fear}. Sensors such as cameras, LiDAR, and 4D radar play a pivotal role by supplying critical data that enables vehicles to make precise, real-time decisions based on a comprehensive understanding of their environment\cite{lin2024rcbevdet,jeong2024spatio,peng2024mufasa,zhang2023photonic}.

In recent years, numerous multimodal datasets, such as KITTI\cite{geiger2013vision}, nuScenes\cite{caesar2020nuscenes}, Waymo\cite{sun2020scalability} and K-Radar\cite{paek2022k}, have emerged in the field of autonomous driving to advance the development of perception methods. However, these datasets typically consist of filtered, clean data, and models trained on them often perform poorly when applied to scenarios involving sensor noise and harsh conditions\cite{dong2023benchmarking}. For instance, under snowy conditions, sensors may be affected by snow accumulation or reflection, resulting in data loss or increased errors. In foggy weather, the sensor’s visibility range may be significantly reduced, affecting the accuracy of object recognition and path planning. Moreover, under intense lighting or low-light conditions at night, cameras and LiDAR may fail to capture sufficient details, undermining the reliability of the decision-making process. According to the U.S. Federal Highway Administration (FHWA), over 5.89 million crashes occur annually. Of these, approximately 21\% (nearly 1.24 million) are related to weather and road conditions, with 42\% occurring on slippery surfaces, 28\% during rainfall, 9\% in snow or sleet, 7\% on icy surfaces, and 8\% on snowy or slushy roads. On average, nearly 5,000 people are killed, and more than 418,000 are injured each year in crashes linked to severe weather and poor road conditions\cite{FHWA}. Sensors are particularly vulnerable to adverse weather and poor road conditions, and existing datasets contain limited samples of sensor anomalies induced by these factors. Therefore, investigating the robustness of autonomous driving perception of sensor anomalies induced by these factors. Therefore, investigating the robustness of autonomous driving perception methods to sensor anomaly data is essential for enhancing the safety of autonomous driving systems\cite{chan2023noise,kamann2020benchmarking,klinghoffer2023towards,zhou2022understanding,zhu2023understanding}.

\begin{figure*}
  \begin{center}
  \includegraphics[width=6.5in]{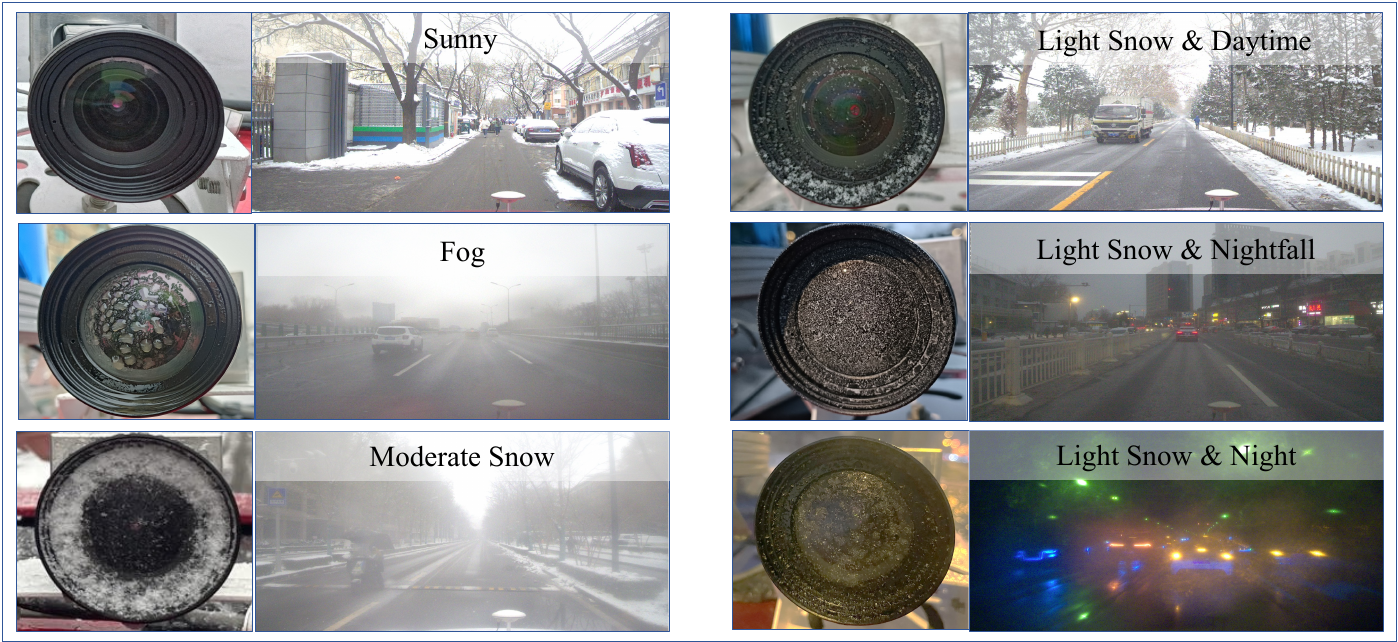}
  \caption{The impact of diverse adverse weather conditions and driving scenarios on camera lens clarity and captured image quality in autonomous driving environments.}\label{Camera_equipment}
  \end{center}
\end{figure*}

Physical simulation is a key approach to meeting the vast data demands of autonomous driving, particularly for rare scenarios. Recently, researchers have employed physical simulations to introduce noise into clean datasets, creating synthetic scenarios that more accurately reflect real-world conditions. Datasets such as KITTI-C\cite{dong2023benchmarking}, Robo3D\cite{kong2023robo3d}, RoboDepth\cite{kong2024robodepth}, and other benchmark datasets have been enhanced by artificially adding simulated noise to help researchers gain a more comprehensive understanding of model performance in complex, noisy environments. These efforts aim to ensure that autonomous driving systems maintain efficient and reliable scenarios through the deliberate corruption of clean data. Noisy datasets for autonomous driving include both image and point cloud noise, reflecting differences in performance under challenging conditions. The development of noisy datasets has significantly advanced the study of model robustness by simulating real-world data formats. Corruption methods to introduce noise include image corruption techniques and point cloud corruption methods. For example, the KITTI-C\cite{dong2023benchmarking} dataset uses the ImgAug image library for image corruption and the Gaussian noise method to corrupt point cloud data. However, differences between these simulated scenarios and the real world still require further validation, with two main limitations: (1) Simulation dataset benchmarks cannot assess model robustness in real-world scenarios. For instance, in snowy conditions, sensors can be obstructed by freezing, water droplets, or snow, and may encounter objects such as cars covered in snow. Existing benchmarks fail to account for these real-world conditions, rendering them ineffective in accurately evaluating algorithm robustness in practical settings, as shown in Fig.\ref{Camera_equipment}. (2) The simulated dataset benchmark has not significantly improved model robustness in real-world environments for certain special scenarios. Since simulated data cannot perfectly replicate all real-world scenes, current simulation methods are insufficient for enhancing the model’s adaptability and robustness in real-world conditions.

To address these issues and fill this gap, this study employs a combination of a camera, LiDAR, and two 4D radar sensors to collect data on common sensor anomalies in extreme weather scenarios. We introduce a benchmark for comparative analysis of autonomous driving performance between simulated and real-world scenarios under adverse weather conditions (S2R-Bench). S2R-Bench covers a wide range of scenarios, including both daytime and nighttime conditions, to comprehensively assess the performance of autonomous driving systems in various environments. It encompasses different road types, including rural, urban, suburban, highways, and tunnels, aiming to simulate diverse driving situations. Additionally, S2R-Bench accounts for various weather conditions, such as light snow, moderate snow, post-snow, and fog, providing a more realistic reflection of how environmental factors impact the performance of autonomous driving systems. We compare simulated sensor anomaly data with real sensor anomaly data to evaluate and enhance the robustness of perception algorithms. To the best of our knowledge, this is the first data benchmark specifically designed to evaluate and improve the robustness of algorithms in real-world scenarios. Our dataset makes a significant contribution to the evaluation of algorithm robustness in real-world scenarios and offers the following benefits: 
\begin{enumerate}

\item We propose a sim-to-real evaluation benchmark for autonomous driving designed to facilitate the comparison and validation between physical simulations and real-world scenarios. This benchmark encompasses various road types, including rural, urban, suburban, highways, and tunnels, allowing for the simulation of diverse driving environments and traffic conditions. Additionally, we consider special weather conditions, such as light snow, moderate snow, post-snow, and fog, which can significantly impact sensor data and the decision-making processes of autonomous driving systems. Our benchmark provides a more comprehensive and realistic assessment of autonomous driving technologies in different scenarios, thereby improving the evaluation and validation of physical simulations against real-world conditions systems. Our benchmarks enable a more comprehensive and realistic assess. 

\item Our benchmark includes camera data, LiDAR data, and two types of 4D radar data, comprising 151 continuous time sequences, each lasting approximately 20 seconds. The data in each sequence is precisely time-synchronized to ensure alignment across sensors at the same time point. In total, 10,117 frames have been meticulously synchronized and processed, ensuring high temporal and spatial accuracy in each frame, thus supporting further analysis and experimentation.

\item We train and validate different approaches using both simulated and real data datasets. The validation confirms the effectiveness of simulation-based methods in real-world applications, demonstrating that the algorithm maintains robust performance and reliability across a wide range of complex environments.
\end{enumerate}

\section{Data Records}

\subsection {Sensor Specification}

\begin{table*}[h!]
    \begin{center}
    \renewcommand\arraystretch{1.1}
    \centering
    \setlength{\tabcolsep}{10pt}
    \caption{The configuration of the autonomous vehicle system platform}
    \resizebox{\textwidth}{!}{
    \begin{tabular}{cccccccccc}
        \toprule[1pt]
            \multirow{2}{*}{{\textbf{Sensors}}}& \multirow{2}{*}{{\textbf{Type}}} & \multicolumn{3}{c}{{\textbf{Resolution}}} & & \multicolumn{3}{c}{{\textbf{FOV}}} & \multirow{2}{*}{{\textbf{FPS}}}  \\
            \cline{3-5} \cline{7-9} & & {\textbf{Range}} & {\textbf{Azimuth}} & {\textbf{Elevation}} & & {\textbf{Range}} & {\textbf{Azimuth}} & {\textbf{Elevation}} &  \\
        \midrule[0.4pt]
        Camera & acA1920-40uc & -- & 1920px & 1200px & & -- & -- & -- & 10 \\
        LiDAR & RS-Ruby Lite & 0.05m & 0.2\textdegree & 0.2\textdegree & & 230m & 360\textdegree & 40\textdegree & 10 \\
        \multirow{2}{*}{4D radar} & Oculli-Eagle  & 0.86m &   \makecell[c]{ \textless 1\textdegree} & \textless 1\textdegree & & 400m & $\pm$56.5\textdegree &  \makecell[c]{ $\pm$22.5\textdegree}& 15 \\
         & Arbe Phoenix & 0.3m & 1.25 \textdegree & 2\textdegree & & 153.6m & 100\textdegree & 30\textdegree & 20 \\
        \bottomrule[1pt]
    \end{tabular}
    }
    
  \label{tab:configuration}
\end{center}
\end{table*}

\begin{figure*}
  \begin{center}
  \includegraphics[width=7in]{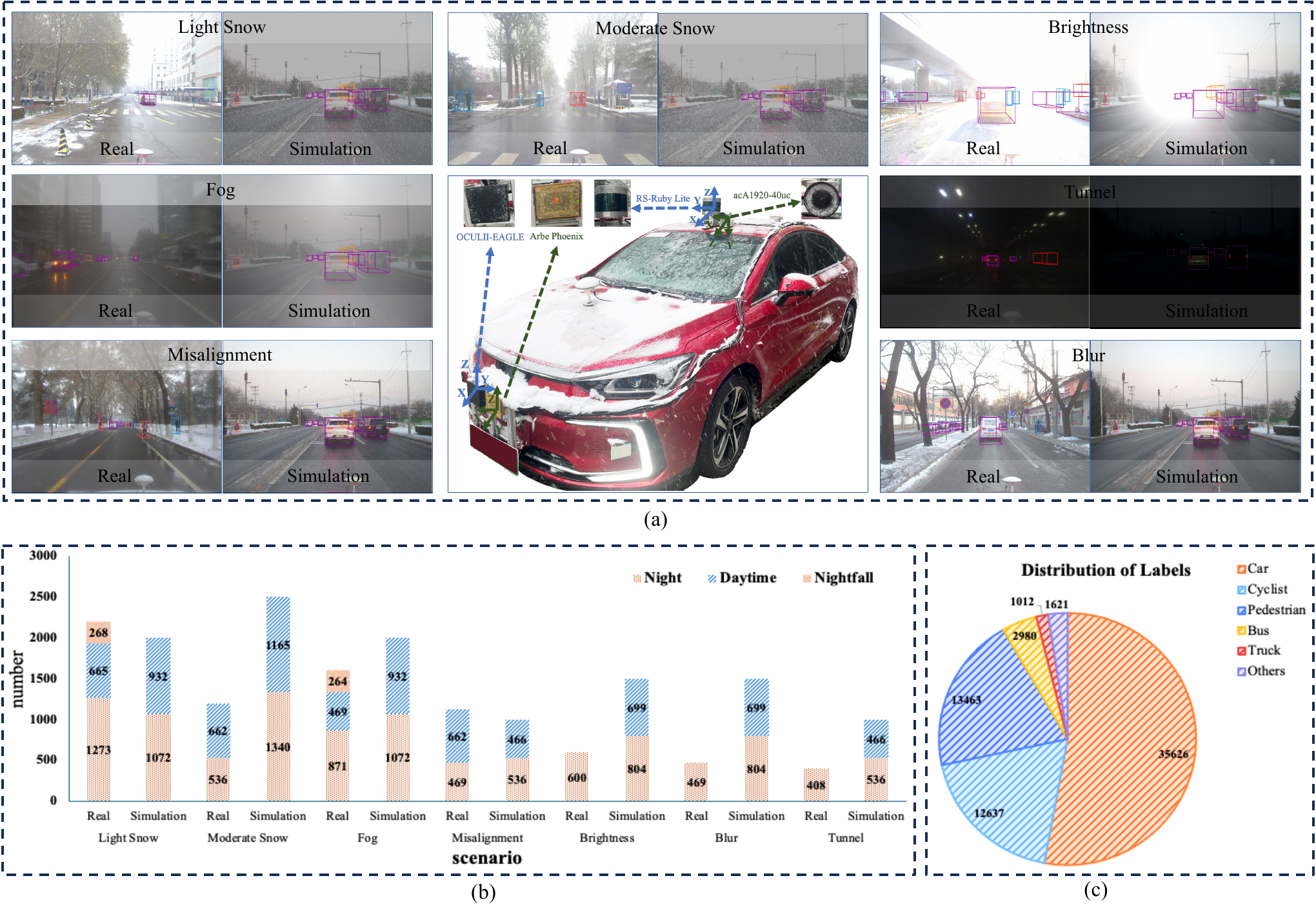}
  \caption{The configuration of our experiment platform and visualisation scenarios. (a) Shows the self-driving car system as well as the real anomaly data collected and the simulated anomaly data. (b) shows the statistics of the number of frames for various periods in different scenarios. (c) shows the statistics of the number of objects in different labels. The result suggests that the main labels like Car, Pedestrian, and Cyclist take up over three-quarters of the total amount of objects.}\label{Autonomous_vehicle_system}
  \end{center}
\end{figure*}

The configuration of our ego vehicle and the coordinate relationships between the multiple sensors are illustrated in Fig. \ref{Autonomous_vehicle_system} (a). The ego vehicle system consists of a high-resolution camera, an 80-line LiDAR, and two types of 4D radar. Sensor configurations are detailed in Table \ref{tab:configuration}. The LiDAR sensor provides an accuracy of 0.2° in both azimuth and elevation, with a 360° horizontal field of view and a 40° vertical field of view. For labeling purposes, we focus on the range in front of the vehicle, retaining data within a 120° field of view. The Oculli-Eagle radar, in long-range mode, captures data within a 113° horizontal and 25° vertical field of view. The Arbe Phoenix radar, in mid-range mode, collects data within a 30° vertical field of view. Detailed sensor specifications are provided in Table \ref{tab:configuration}.

\subsection {Data collection}

Our benchmark was conducted from December 13, 2023, to January 18, 2024, in Beijing, China. We specifically selected a variety of road conditions, including city roads, suburban roads, highways, tunnels, towns, villages, communities, and campuses, covering approximately 700 km. Additionally, data was collected during both daytime and nighttime under various weather conditions, including light snow, moderate snow, post-snow, sunny, cloudy, foggy, and tunnel scenarios, ensuring data diversity and providing a benchmark for the in-depth study of sensor anomalies in real-world environments.
\subsection {Data processing}
In this study, we developed a multi-sensor data acquisition system comprising a Basler aca1920-40gc camera, an 80-line LiDAR, and two types of 4D radar, augmented with RTK for precise raw time signal acquisition. The data streams from each sensor are collected through separate processes within the ROS framework. To calibrate both the internal and external parameters of the camera, we applied Zhang Zhengyou’s calibration\cite{zhengyou1998flexible} method, achieving accurate results by capturing images of a checkerboard grid from various viewpoints. Time synchronization between sensors is critical for autonomous driving, and we used the auto66 time synchronization box from CoolShark Technology Inc to ensure precise timing. The camera data was captured using a 10Hz PPS signal trigger, while the two 4D radars and the LiDAR were synchronized via the PTP signal from the synchronization box. This integrated data acquisition approach significantly improves data processing efficiency, enabling the autonomous driving system to rely on real-time, accurate sensor data for environmental sensing and decision-making. By precisely synchronizing the timestamps across sensors, our system provides robust data support for complex autonomous driving applications, establishing a strong foundation for future research and practical implementations.

\begin{figure*}
  \begin{center}
  \includegraphics[width=6in]{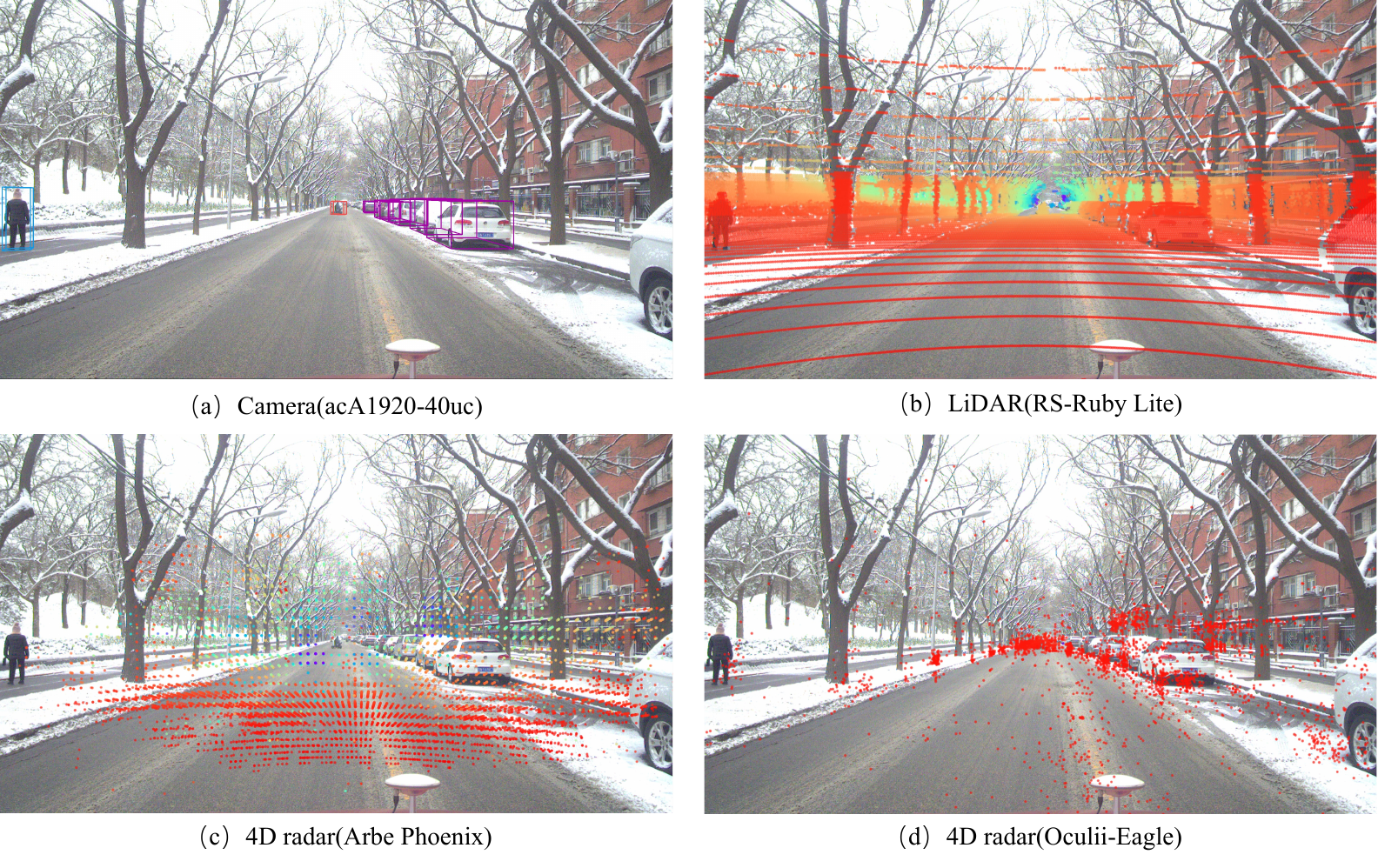}
  \caption{Projection visualization of sensor calibration. (a), (b), (c), and (d) represent the projection of the calibrated data (3D bounding box, LiDAR point cloud, Arbe Phoenix point cloud, and Oculli-Eagle point cloud) on the image.}\label{Data_calibration}
  \end{center}
\end{figure*}

The calibration process for this dataset is divided into two main steps: joint camera-LiDAR calibration and joint camera-4D radar calibration, both of which are conducted ofﬂine. Traditional methods for calibrating multi-modal sensors, particularly camera, both of which are performed offline. Traditional methods for calibrating multimodal sensors, particularly cameras and LiDAR\cite{geiger2012automatic}, have been well-established in previous studies. Existing 3D radar calibration techniques, which are based on LiDAR calibration, have also proven successful due to the similarities between the characteristics of 3D and 4D radars\cite{el2015radar}. Therefore, the method used to obtain the internal and external parameters of the 4D radar follows the approach employed in 3D radar calibration. The calibration results are shown in Fig. \ref{Data_calibration}.

In our ego vehicle system, the origin of the LiDAR coordinate system serves as the reference for the multi-sensor relative coordinate system. Offline calibration is used to obtain precise calibration results, offering flexibility in both site selection and calibration methods\cite{dhall2017LiDAR,pervsic2019extrinsic}. We select a flat field with normal lighting on a sunny day for the camera calibration. A rigid calibration plate is positioned in front of the sensor system, and a spherical coordinate system is employed to extract and compute the LiDAR’s external parameters using the principle of rigid transformation and the 3D data of the sphere. Joint camera-4D radar calibration is carried out with tools such as a corner reflector and a calibration plate. The sensitivity of the 4D radar to metal points is utilized to calculate and extract the radar camera-4D radar calibration is calibration parameters. Through this process, joint calibration of the camera, LiDAR, and 4D radar is successfully completed.

\subsection {Dataset organisation}
\begin{figure*}[h!]
  \begin{center}
  \includegraphics[width=7in]{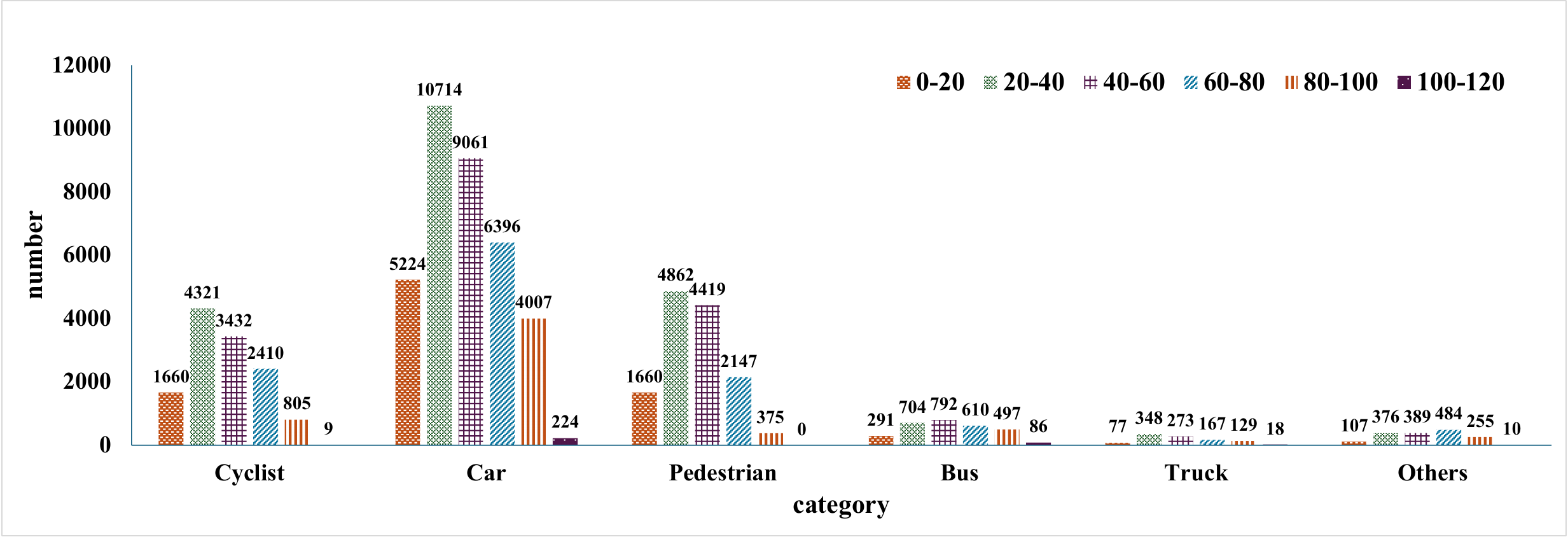}
  \caption{The statistic of different annotated objects at \textbf{different ranges of distance} from the ego vehicle. From the results, the majority of the annotated objects are in the range of 20m-80m.}\label{Distance_range}
  \end{center}
\end{figure*}

Our benchmark comprises three components: a real sensor anomaly dataset (\textcolor{cyan}{S2R-R}), a clean dataset (\textcolor{red}{S2R-C}), and a simulated sensor snomaly dataset (\textcolor{blue}{S2R-S}), with a total of 151 sequences (10,117 frames) labeled. Each sequence is approximately 20 seconds in duration. The \textcolor{cyan}{S2R-R} is primarily collected under severe weather conditions, while the \textcolor{red}{S2R-C} is collected under normal weather conditions. The (\textcolor{blue}{S2R-S}) consists of simulated data for each \textcolor{cyan}{S2R-R} scenario, generated using five commonly used simulation methods. This simulation includes four sequences representing both daytime and nighttime conditions in the \textcolor{red}{S2R-C}. The calibration and annotation the simulated dataset align with those of the 8 sequences in the \textcolor{red}{S2R-C}. Ultimately, the dataset will be released in KITTI format. A statistical analysis was conducted on the 10,117 labeled frames, counting the number of objects in each label category, as shown in Fig. \ref{Autonomous_vehicle_system}(c). A pie chart illustrates the distribution of objects across the six most frequent label categories, with the “others” category predominantly consisting of tricycles. As shown in Fig. \ref{Autonomous_vehicle_system}(c), the majority of labels are concentrated in the categories of “car”, “pedestrian” and “cyclist”, accounting for approximately 53\%, 20\%, and 19\%, respectively. Additionally, we analyzed the number of objects within different distance ranges. As shown in Fig. \ref{Distance_range}, most objects are located within 80 meters of the ego vehicle.

We also quantified the number of frames in each scenario of the dataset, as shown in Fig. \ref{Autonomous_vehicle_system}(b). The real-world data is approximately equal in volume to the simulated data. About three-quarters of the real-world data is attributed to the \textcolor{cyan}{S2R-R}, while the remaining one quarter corresponds to the \textcolor{red}{S2R-C}. Data collected during the daytime, nighttime, and dusk accounted for 51.3\%, 41.3\%, and 7.4\% of the real-world dataset, respectively. Data collected in tunnels is classified as density anomaly data, while data exhibiting spatial asynchrony between the point cloud and image information—due to uneven roads or speed bumps during driving—is classified as spatial misalignment data. Data with blurred images, caused by the formation of water mist or dust due to significant indoor/outdoor temperature differences on snowy days, is classified as fuzzy data. We collected 2,206 frames in light snow, 1,198 frames in moderate snow, 1,604 frames in fog, and 600 frames under strong lighting or exposure conditions, which account for approximately 27.3\%, 14.9\%, 19.9\%, and 7.4\% of the \textcolor{cyan}{S2R-R}, respectively. 

The data collected under severe weather conditions is essential for evaluating the performance of 4D radar sensors in these environments, which is crucial for advancing perception algorithms based on 4D radar point clouds. Additionally, this data can help assess whether different sensing methods exhibit consistent robustness in specific scenarios, a key consideration for the field of autonomous driving.

\subsection {Dataset directory}

Our benchmark is available in both data repositories \cite{WangLi_dataset1,WangLi_dataset2,WangLi_dataset3, WangLi_dataset4, WangLi_dataset5}. The dataset consists of three parts: \textcolor{red}{S2R-C}, \textcolor{cyan}{S2R-R}, and \textcolor{blue}{S2R-S}. First, \textcolor{red}{S2R-C} is divided into training, validation, and test sets in an 8:1:1 ratio. \textcolor{cyan}{S2R-R} is divided into training, validation, and test sets in the proportions of 52.3\%, 25.3\%, and 22.4\%, respectively. The \textcolor{blue}{S2R-S} dataset is divided into training, validation, and test sets in the proportions of 52.1\%, 29.2\%, and 23.7\%, respectively. Additionally, the \textcolor{blue}{S2R-S} dataset is generated using four simulation methods: 3D$\_$Corruptions$\_$AD\cite{dong2023benchmarking}, MultiCorrupt\cite{beemelmanns2024multicorrupt}, Robo3D\cite{kong2023robo3d}, and RoboDepth\cite{kong2024robodepth}, and contains four subsets: “\textcolor{blue}{\textcolor{blue}{S2R-S}-3D$\_$Corruptions$\_$AD}\cite{dong2023benchmarking}”, “\textcolor{blue}{S2R-S-MultiCorrupt}\cite{beemelmanns2024multicorrupt}”, “\textcolor{blue}{S2R-S-Robo3D}\cite{kong2023robo3d}” and “\textcolor{blue}{S2R-S-Robodepth}\cite{kong2024robodepth}”. Note that the RoboDepth simulation method can only simulate image data.

\subsection {Data Visualization}

\begin{figure*}[h!]
  \begin{center}
  \includegraphics[width=7in]{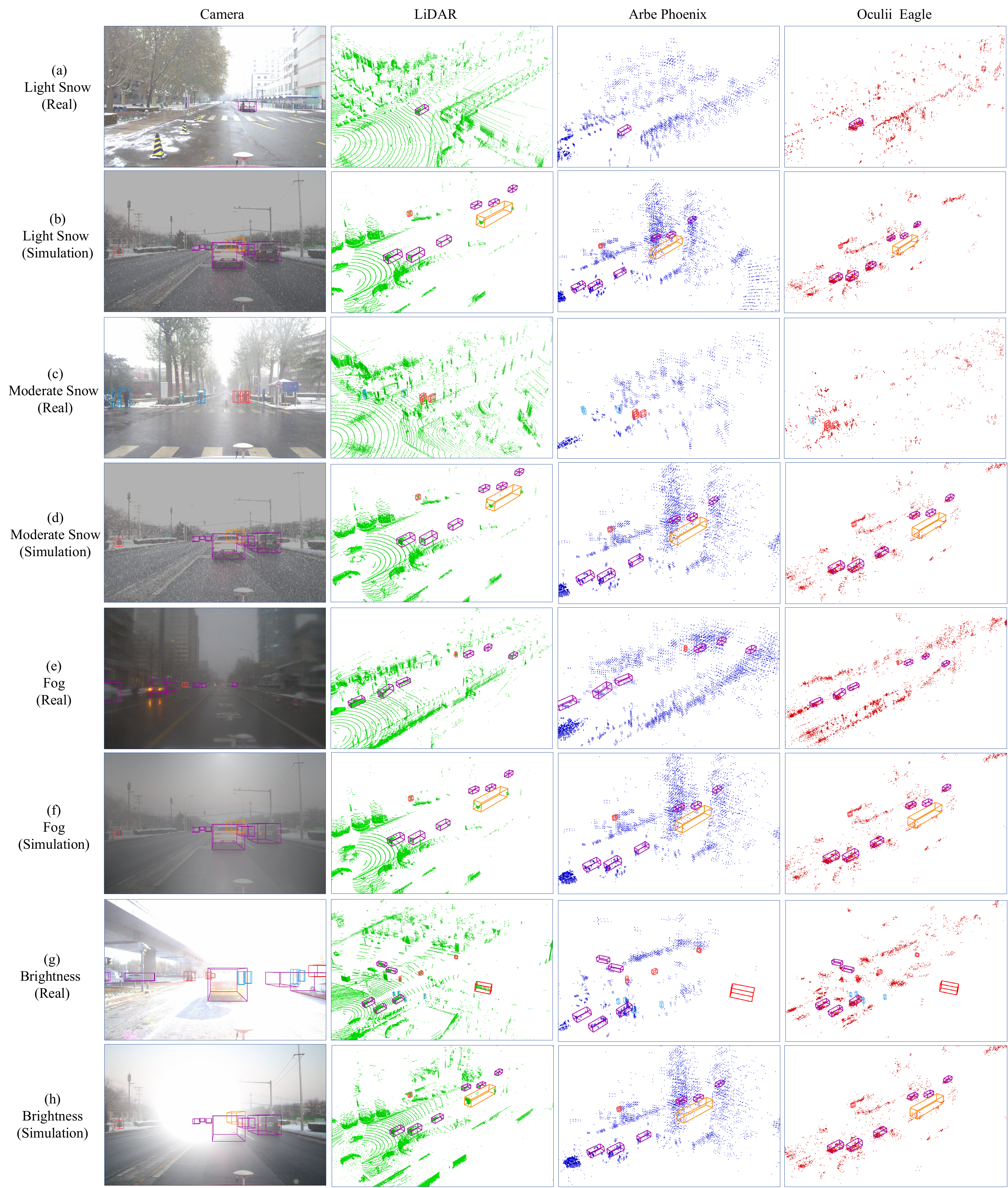}
  \caption{Representing 3D annotations in multiple scenarios and sensor modalities. The four columns respectively display the projection of 3D annotation boxes in images, LiDAR point clouds, Arbe Phoenix and Oculli-Eagle radar point clouds. Each row represents a scenario type. (a) light snow $\&$ real; (b) light snow $\&$ simulation; (c) moderate snow $\&$ real; (d) moderate snow $\&$ simulation; (e) fog $\&$ real; (f) fog $\&$ simulation; (g) brightness $\&$ real; (h) brightness $\&$ simulation.}\label{Fig_visualization1}
  \end{center}
\end{figure*} 
\begin{figure*}[h!]
  \begin{center}
  \includegraphics[width=7in]{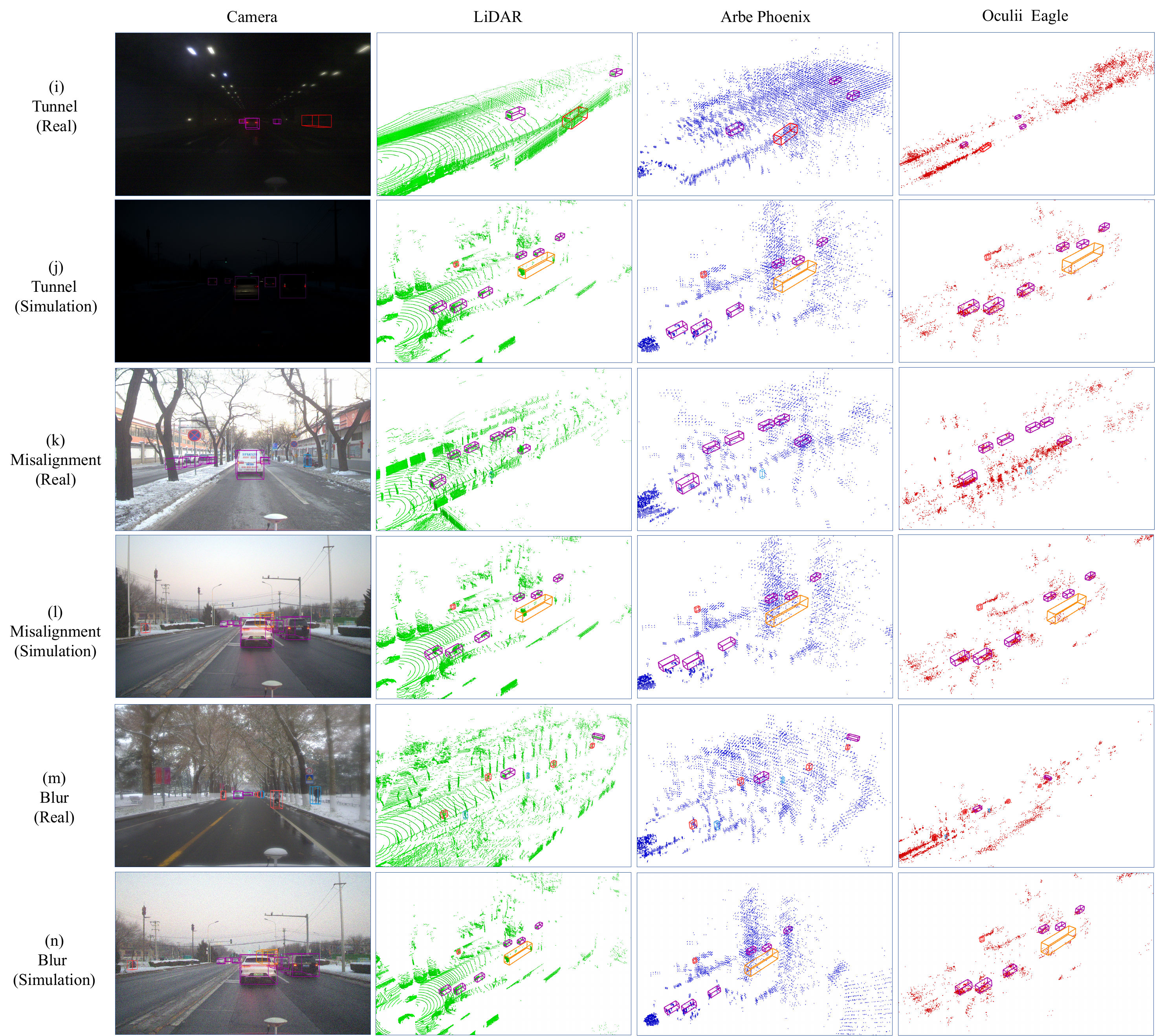}
  \caption{Representing 3D annotations in multiple scenarios and sensor modalities. Each row represents one scenario. (i) tunnel $\&$ real; (j) tunnel $\&$ simulation; (k) misalignment $\&$ real; (l) misalignment $\&$ simulation; (m) blur $\&$ real; (n) blur $\&$ simulation. }\label{Fig_visualization2}
  \end{center}
\end{figure*} 

As shown in Fig. \ref{Fig_visualization1} and \ref{Fig_visualization2}, we visualized the data from the \textcolor{cyan}{S2R-R} dataset alongside data from corresponding scenarios simulated using the 3D$\_$Corruptions$\_$AD method to compare real-world and simulated data. Objects are labeled with 3D bounding boxes, which are then projected onto the image, LiDAR, and two 4D radar point clouds. The figures demonstrate that the 3D bounding boxes accurately fit the objects and align well with the corresponding points in the point clouds. Object in both the LiDAR and 4D radar point clouds correspond to those in the image, further confirming the synchronization between the sensors. The odd-numbered rows represent visualizations from real-world scenarios (e.g., Fig. \ref{Fig_visualization1} (a, c, e, g) and Fig. \ref{Fig_visualization2} (i, k, m)), which are affected by weather and lighting conditions, resulting in incomplete or unclear RGB information captured by the camera. This limitation is compensated for by the LiDAR and 4D radar. In snowy scenarios, LiDAR is particularly prone to interference from snowflakes, introducing significant noise, while the 4D radar is less affected by weather and provides more accurate object characterization. The even-numbered rows correspond to simulated data visualizations (e.g., Fig. \ref{Fig_visualization1} (b, d, f, h) and Fig. \ref{Fig_visualization2} (j, l, n)), which show noticeable differences characterization. The even-numbered rows correspond to simulated data visualizations (e.g., Fig. \ref{Fig_visualization1} (b, d, f, h) and Fig. \ref{Fig_visualization2} (j, l, n)), which show noticeable differences compared to the real-world data.

\section {Methods }
\subsection {Simulation Methods}

We introduce the simulation methods used for each scenario and explain how they enable the replication of various real-world situations, thereby supporting subsequent analysis and experiments.
\subsubsection {Snow}
In snowy conditions, snow particles in the air absorb and scatter the laser beam, while snowflake reflections introduce significant noise in the LiDAR point cloud\cite{hahner2022LiDAR,sun2024understanding}. This results in anomalies in the received power of the point cloud, leading to erroneous signals from targets at varying distances. Additionally, snow particles can severely obscure the target in camera data, making it difficult for both humans and autonomous driving algorithms to make accurate judgments. To simulate snowy conditions on clean data, we use the 3D$\_$Corruptions$\_$AD algorithm\cite{dong2023benchmarking}, MultiCorrupt\cite{beemelmanns2024multicorrupt}, Robo3D\cite{kong2023robo3d}, and RoboDepth\cite{kong2024robodepth} algorithms.

\subsubsection {Fog}
The fog scenario differs from rainy and snowy conditions in that it exhibits an uneven distribution, meaning it is not uniformly spread around the sensors\cite{shi2022research,li2023domain}. We adopt the method used in Robo3D, where water droplets forming the fog are modeled as transparent spheres and placed irregularly around the LiDAR and radar sensors. To simulate fog noise in images, we employ the fog noise simulation function from the imgaug library.

\subsubsection {Brightness}

During daytime LiDAR and 4D radar measurements, sunlight contamination primarily contributes to backscatter noise, with the detector receiving both the laser return signal and background sunlight\cite{zhang2023perception}. These photons, despite originating from different sources, converge at the same position on the sensor because the plane wave from the ambient sunlight and the Gaussian beam, when reflected and received by the sensor’s optics do not differ significantly. Specifically, under bright light conditions, the increased reflection of scattered sunlight from surface (excluding the ground) can substantially interfere with the intensity of the pulses returned to the sensor. Additionally, glare and lens flare in images can cause interference, resulting in degraded image quality. To simulate lighting anomalies in clean data, we employ the 3D$\_$Corruptions$\_$AD\cite{dong2023benchmarking}, MultiCorrupt\cite{beemelmanns2024multicorrupt} methods.

\subsubsection {Spatial Misalignment}

In real-world scenarios, road bumps and errors in sensor calibration often result in spatial offsets between the point cloud and the camera input. These offsets can significantly impact the fusion quality of multimodal approaches\cite{guindel2017automatic}. To simulate such spatial offsets, we use the 3D$\_$Corruptions$\_$AD\cite{dong2023benchmarking} and MultiCorrupt\cite{beemelmanns2024multicorrupt} methods to apply translation and rotation noise to the point cloud, while keeping the projection matrix unchanged.

\subsubsection {Blurred}

During the processes of capturing, encoding, and transmitting images, thermal effects, signal interference, and compression artififacts can introduce noise, degrading both image quality and clarity\cite{li2021automatic,secci2020failures}. Gaussian noise, a common model for randomness and uncertainty, is used to simulate real-world noise. To simulate image blurring on clean data, we employ the 3D$\_$Corruptions$\_$AD\cite{dong2023benchmarking} and MultiCorrupt\cite{beemelmanns2024multicorrupt} methods.
\subsubsection {Tunnel}

In tunnel scenarios, anomalies in point cloud density can introduce additional clutter and noise, which interfere with the contours and shapes of actual objects, thereby degrading the overall quality of the point cloud\cite{lin2021automatic,heidecker2021application}. Moreover, high-density point clouds necessitate more sophisticated processing and filtering algorithms to remove noise and extraneous points, facilitating the extraction of useful information and features. To simulate density anomaly scenarios on clean data, we employ the 3D$\_$Corruptions$\_$AD\cite{dong2023benchmarking}, MultiCorrupt\cite{beemelmanns2024multicorrupt} methods.

\subsection {Data annotation}

Due to the differing data frequencies of the camera, LiDAR, and both 4D radars, we selected synchronized frames from the collected data. To align timestamps across the sensors, we utilized the Precision Time Protocol (PTP), which leverages GPS timing and synchronization devices to adjust the time between multiple sensors. For the synchronized frames, we provided annotation details such as 3D bounding boxes, object labels, and tracking IDs for each object. The annotations were based on the LiDAR point cloud data, with corresponding information for the other sensors derived by transforming the LiDAR data into each sensor's external reference matrix. For each object within a 100-meter range, we provided key details including transforming the LiDAR data into each sensor's external reference matrix. For each object within a 100-meter range, we provided key details including the relative coordinates (\emph{x}, \emph{y}, \emph{z}) of the 3D bounding box, absolute dimensions (length, width, height), and the orientation angle (alpha) in the Bird's Eye View (BEV), all of which required precise calculations. To ensure data usability, we also included features indicating object occlusion and truncations. We annotated over a dozen categories, with a focus on five key categories: “car”, “pedestrian”, “cyclist”, “bus”, “truck” and “other”.

\subsection {Methods for technical validation}

We employed a server running Ubuntu 18.04 as the hardware platform for our experiments. The OpenPCDet project, based on PyTorch 10.2, was adopted, with a batch size set to the default value of 4. Each experiment was trained for 80 epochs with a learning rate of 0.003 on four Nvidia RTX 3090 graphics cards. To evaluate the performance of our dataset and the designed algorithms, we compared them with several state-of-the-art algorithms, all of which were tested as required. Both unimodal and multimodal baseline models were used, including PointPillars\cite{lang2019pointpillars}, SMOKE\cite{liu2020smoke}, and the multimodal baseline model Focals Conv\cite{chen2022focal}.

\section{EXPERIMENT}
This section outlines the experimental platform used to evaluate our dataset. We employ the PointPillars and SMOKE baselines to assess its performance. The experimental results confirm the dataset's effectiveness. We then analyze the results both qualitatively and quantitatively, providing a comprehensive evaluation of the dataset. We define cases where the data variation is within 30\% as indicating a good simulation effect. Within this range, the simulation results are considered highly accurate and closely reflective of real-world scenarios, thus enhancing the reliability and validity of the outcomes. These cases are highlighted in bold in the experimental results table.

\subsection {Baseline model comparison on a S2R-C.}

\begin{table*}[]

    \begin{center}
        \renewcommand\arraystretch{1.1}
        \centering
        \setlength{\tabcolsep}{10pt}
        \caption{The comparison of the “Car” category with the Pointpillars model at medium difficulty using “\textcolor{red}{S2R-C}” for training, the “\textcolor{cyan}{S2R-R}” and “\textcolor{blue}{S2R-S}” were used for testing. “\textcolor{red}{S2R-C}” represents the clean and clear dataset collected without anomalous noise; “\textcolor{cyan}{S2R-R}” represents the real sensor anomaly test dataset collected; “\textcolor{blue}{S2R-S}” represents a test dataset of sensor anomaly simulated using simulation methods. “\textcolor{blue}{S2R-S}” includes three subsets: “\textcolor{blue}{\textcolor{blue}{\textcolor{blue}{S2R-S}-3D$\_$Corruptions$\_$AD}}\cite{dong2023benchmarking}”, “\textcolor{blue}{\textcolor{blue}{\textcolor{blue}{\textcolor{blue}{\textcolor{blue}{S2R-S}-MultiCorrupt}}}}\cite{beemelmanns2024multicorrupt}”,  and “\textcolor{blue}{\textcolor{blue}{S2R-S}-Robo3D}\cite{kong2023robo3d}”. The data variation within 30\% is defined as good simulation performance and is highlighted in \textbf{bold}.}
        
    \resizebox{\textwidth}{!}{
    \begin{tabular}{ccccccccccccccc}
        \toprule[1pt]
        \multicolumn{2}{c}{Dataset}                               & \multicolumn{1}{c}{\multirow{2}{*}{Type}} & \multicolumn{2}{c}{Light Snow}                   & \multicolumn{2}{c}{Moderate Snow}                & \multicolumn{2}{c}{Fog}                          & \multicolumn{2}{c}{Brightness}                         & \multicolumn{2}{c}{Tunnel}       & \multicolumn{2}{c}{Misalignment}                  \\
        \cline{4-15}
        
        \multicolumn{1}{c}{Train} & \multicolumn{1}{c}{Test}      & \multicolumn{1}{c}{}                      & \multicolumn{1}{c}{3D} & \multicolumn{1}{c}{BEV} & \multicolumn{1}{c}{3D} & \multicolumn{1}{c}{BEV} & \multicolumn{1}{c}{3D} & \multicolumn{1}{c}{BEV} & 3D                        & BEV                          & 3D                        & BEV   & 3D                    & BEV                   \\
        
\midrule[0.4pt]

\multirow{12}{*}{\textcolor{red}{S2R-C}}   & \multirow{3}{*}{\textcolor{cyan}{S2R-R}}                 & Lidar                 & 54.05          & 54.06          & 58.23           & 58.23           & 37.63      & 50.41      & 31.07          & 31.13          & 67.39        & 67.39        & 40.19         & 41.33        \\
&  & Arbe                  & 32.57          & 32.61          & 44.05           & 44.63           & 8.99       & 10.81      & 17.11          & 17.73          & 60.88        & 61.05        & 27.16         & 28.64        \\
&  & Occuli                & 25.09          & 25.40          & 16.94           & 19.47           & 2.73       & 5.68       & 11.01          & 12.13          & 43.76        & 50.79        & 10.93         & 13.89        \\
\cline{2-15}
 & \multirow{3}{*}{\makecell[c]{\textcolor{blue}{S2R-S}\\ \textcolor{blue}{-3D$\_$Corruptions$\_$AD}\cite{dong2023benchmarking}}}                  & Lidar                 & 25.59          & 25.79          & 22.74           & 23.05           & 24.42      & 25.17      & \textbf{24.59} & \textbf{24.59} & 34.37        & 34.88        & 3.72          & 23.60        \\
 & & Arbe                  & \textbf{29.36} & \textbf{29.48} & 28.44           & 28.95           & 28.58      & 29.07      & 1.54           & 1.80           & 27.37        & 28.20        & 0.00          & 0.00         \\
 & & Occuli                & \textbf{21.03} & \textbf{21.59} & \textbf{21.64}  & \textbf{22.86}  & 13.50      & 18.19      & 1.78           & 2.23           & 21.50        & 22.50        & 0.00          & 0.00         \\
\cline{2-15}
 & \multirow{3}{*}{\textcolor{blue}{\textcolor{blue}{\textcolor{blue}{\textcolor{blue}{\textcolor{blue}{S2R-S}-MultiCorrupt}}}}\cite{beemelmanns2024multicorrupt}}        & Lidar                 & 33.20          & 34.34          & 33.20           & 34.34           & 24.42      & 25.17      & \textbf{33.20} & \textbf{34.34} & 33.92        & 33.92        & 6.78          & 7.80         \\
 & & Arbe                  & \textbf{28.89} & \textbf{29.24} & 28.89           & 29.24           & 28.58      & 29.07      & 28.89          & 29.24          & 18.17        & 22.00        & 0.00          & 0.00         \\
 & & Occuli                & \textbf{21.77} & \textbf{22.46} & \textbf{21.77}  & \textbf{22.46}  & 13.50      & 18.19      & 1.78           & 2.23           & 21.56        & 23.21        & 0.00          & 0.00         \\
\cline{2-15}
 & \multirow{3}{*}{\textcolor{blue}{\textcolor{blue}{S2R-S}-Robo3D}\cite{kong2023robo3d}}              & Lidar                 & 15.09          & 16.27          & 18.58           & 20.60           & 24.42      & 25.17      & \textbf{33.20} & \textbf{34.34} & 33.20        & 34.34        & 5.23          & 7.02         \\
 & & Arbe                  & 13.97          & 17.41          & 13.77           & 16.39           & 28.58      & 29.07      & 24.92          & 26.23          & 28.89        & 29.24        & 0.00          & 0.00         \\
 & & Occuli                & 2.70           & 3.81           & 3.76            & 4.61            & 13.50      & 18.19      & 1.78           & 2.23           & 21.77        & 22.46        & 0.00          & 0.00        \\
\bottomrule[1px]
    \end{tabular}
    }
      \label{tab:experiment2}
    \end{center}
    \end{table*}

    \begin{table*}[]
        \begin{center}
            \renewcommand\arraystretch{1.1}
            \centering
            \setlength{\tabcolsep}{10pt}
            \caption{The comparison of the “Car” category with the SMOKE model at medium difficulty using “\textcolor{red}{S2R-C}” for training, the “\textcolor{cyan}{S2R-R}” and “\textcolor{blue}{S2R-S}” were used for testing. “\textcolor{red}{S2R-C}” represents the clean and clear dataset collected without anomalous noise; “\textcolor{cyan}{S2R-R}” represents the real sensor anomaly test dataset collected; “\textcolor{blue}{S2R-S}” represents a test dataset of sensor anomaly simulated using simulation methods. “\textcolor{blue}{S2R-S}” includes four subsets: “\textcolor{blue}{\textcolor{blue}{\textcolor{blue}{S2R-S}-3D$\_$Corruptions$\_$AD}}\cite{dong2023benchmarking}”, “\textcolor{blue}{\textcolor{blue}{\textcolor{blue}{\textcolor{blue}{\textcolor{blue}{S2R-S}-MultiCorrupt}}}}\cite{beemelmanns2024multicorrupt}”, “\textcolor{blue}{\textcolor{blue}{S2R-S}-Robo3D}\cite{kong2023robo3d}” and “\textcolor{blue}{\textcolor{blue}{S2R-S}-Robodepth}\cite{kong2024robodepth}”. The data variation within 30\% is defined as good simulation performance and is highlighted in \textbf{bold}.}
        
        \resizebox{\textwidth}{!}{
        
        \begin{tabular}{cclcccccccccccc}
        \toprule[1pt]
        \multicolumn{2}{c}{Dataset}                       & \multicolumn{1}{c}{}                       & \multicolumn{2}{c}{Light Snow} & \multicolumn{2}{c}{Moderate Snow} & \multicolumn{2}{c}{Fog} & \multicolumn{2}{c}{Brightness} & \multicolumn{2}{c}{Misalignment} & \multicolumn{2}{c}{Image Blurred} \\
        \cline{4-15} 
        Train               & Test                & \multicolumn{1}{c}{\multirow{-2}{*}{Type}} & 3D             & BEV                    & 3D             & BEV           & 3D          & BEV        & 3D                       & BEV                      & 3D            & BEV          & 3D                          & BEV                         \\
        
        \midrule[0.4pt]

        & \textcolor{cyan}{S2R-R}           & Image & 25.22                    & 26.72                        & 9.09                     & 9.09                         & 7.9                      & 8.24                         & 22.44                    & 22.93                        & 15.77                              & 21.03                        & 14.98                    & 15.45                        \\
        & \makecell[c]{\textcolor{blue}{S2R-S}\\ \textcolor{blue}{-3D$\_$Corruptions$\_$AD}\cite{dong2023benchmarking}}           & Image & \textbf{21.81}           & \textbf{22.06}               & 15.35                    & 16.41                        & 29.89                    & 31.13                        & 15.5                     & 15.88                        & 38.78                              & 39.53                        & 23.26                    & 24.06                        \\
        & \textcolor{blue}{\textcolor{blue}{\textcolor{blue}{\textcolor{blue}{\textcolor{blue}{S2R-S}-MultiCorrupt}}}}\cite{beemelmanns2024multicorrupt} & Image & \textbf{31.42}           & \textbf{32.39}               & 22.38                    & 23.93                        & 22.69                    & 24.11                        & \textbf{16.51}           & \textbf{24.24}               & 38.78                              & 39.53                        & 29.63                    & 31.28                        \\
        & \textcolor{blue}{\textcolor{blue}{S2R-S}-Robo3D}\cite{kong2023robo3d}        & Image & 38.78                    & 39.53                        & 38.78                    & 39.53                        & 38.78                    & 39.53                        & 38.78                    & 39.53                        & 38.78                              & 39.53                        & 38.78                    & 39.53                        \\
        \multirow{-5}{*}{\textcolor{red}{S2R-C}} & \textcolor{blue}{\textcolor{blue}{S2R-S}-Robodepth}\cite{kong2024robodepth}     & Image & \textbf{23.06}           & \textbf{23.78}               & 16.41                    & 16.81                        & 30.59                    & 31.31                        & 31.45                    & 32.55                        & 38.78                              & 39.53                        & 28.57                    & 30.08                       \\

        \bottomrule[1px]
        \end{tabular}
        }  
         \label{tab:experiment3}
    \end{center}
    \end{table*}

    \begin{table*}[]
        \renewcommand\arraystretch{1.1}
        \centering
        \setlength{\tabcolsep}{10pt}
        \caption{The comparison of the “Car” category with the Focals Conv model at medium difficulty using \textcolor{red}{S2R-C} for training and Simulated-dataset for testing. “\textcolor{cyan}{S2R-R}” represents the real sensor anomaly test dataset collected; “\textcolor{blue}{S2R-S}” represents a test dataset of sensor anomaly simulated using simulation methods. “\textcolor{blue}{S2R-S}” includes three subsets: “\textcolor{blue}{\textcolor{blue}{\textcolor{blue}{S2R-S}-3D$\_$Corruptions$\_$AD}}\cite{dong2023benchmarking}”, “\textcolor{blue}{\textcolor{blue}{\textcolor{blue}{\textcolor{blue}{\textcolor{blue}{S2R-S}-MultiCorrupt}}}}\cite{beemelmanns2024multicorrupt}”, “\textcolor{blue}{\textcolor{blue}{S2R-S}-Robo3D}\cite{kong2023robo3d}” and “\textcolor{blue}{\textcolor{blue}{S2R-S}-Robodepth}\cite{kong2024robodepth}”. The data variation within 30\% is defined as good simulation performance and is highlighted in \textbf{bold}.}
    
    \resizebox{\textwidth}{!}{
    \begin{tabular}{cclcccccccccccc}
    \toprule[1pt]
    \multicolumn{2}{c}{Dataset}                                    & \multicolumn{1}{c}{\multirow{2}{*}{Type}} & \multicolumn{2}{c}{Light Snow}  & \multicolumn{2}{c}{Moderate Snow} & \multicolumn{2}{c}{Fog}         & \multicolumn{2}{c}{Brightness}  & \multicolumn{2}{c}{Tunnel}     & \multicolumn{2}{c}{Misalignment}    \\
    \cline{4-15}
    Train                          & Test                          & \multicolumn{1}{c}{}                      & 3D             & BEV            & 3D              & BEV             & 3D             & BEV            & 3D             & BEV            & 3D             & BEV            & 3D             & BEV            \\
    \midrule[0.4pt]               
                                     
\multirow{12}{*}{\textcolor{red}{S2R-C}} & \multirow{3}{*}{\textcolor{cyan}{S2R-R}}         & Lidar                                      & 44.21          & 44.23          & 59.64           & 59.64           & 37.63      & 50.41      & 32.04          & 32.06          & 1.78    & 2.23  & 67.21          & 67.21          \\
                                      & & Arbe                                       & 34.36          & 34.78          & 33.09           & 33.58           & 13.98      & 15.08      & 14.55          & 15.22          & 0.55    & 0.85  & 51.97          & 53.35          \\
                                      & & Occuli                                     & 31.04          & 31.41          & 25.08           & 25.09           & 16.25      & 16.49      & 9.87           & 11.35          & 22.32   & 23.82 & 50.79          & 51.47          \\
\cline{2-15}
& \multirow{3}{*}{\makecell[c]{\textcolor{blue}{S2R-S}\\ \textcolor{blue}{-3D$\_$Corruptions$\_$AD}\cite{dong2023benchmarking}}}   & Lidar                                      & 28.07          & 28.07          & 25.95           & 25.95           & 25.61      & 25.61      & \textbf{31.29} & \textbf{31.29} & 25.44   & 25.44 & 34.08          & 34.08          \\
                                      & & Arbe                                       & \textbf{24.14} & \textbf{25.07} & \textbf{25.30}  & \textbf{25.60}  & 25.10      & 25.26      & \textbf{16.74} & \textbf{16.77} & 23.45   & 24.59 & 24.91          & 26.00          \\
                                      & & Occuli                                     & \textbf{24.36} & \textbf{25.44} & \textbf{24.82}  & \textbf{25.54}  & 22.58      & 25.20      & 14.28          & 14.78          & 24.52   & 25.56 & 22.55          & \textbf{25.33} \\
\cline{2-15}
& \multirow{3}{*}{\textcolor{blue}{\textcolor{blue}{\textcolor{blue}{\textcolor{blue}{\textcolor{blue}{S2R-S}-MultiCorrupt}}}}\cite{beemelmanns2024multicorrupt}}                                     & Lidar                                      & \textbf{34.09} & \textbf{34.09} & 34.10           & 34.10           & 25.60      & 25.64      & \textbf{34.15} & \textbf{34.15} & 34.09   & 34.09 & 32.00          & 32.04          \\
                                      & & Arbe                                       & \textbf{25.84} & \textbf{25.78} & \textbf{25.76}  & \textbf{25.88}  & 25.23      & 25.37      & 25.85          & 26.05          & 22.51   & 22.52 & 14.19          & 14.47          \\
                                      & & Occuli                                     & \textbf{24.58} & \textbf{25.73} & \textbf{22.47}  & \textbf{25.46}  & 22.56      & 25.24      & 24.38          & 24.98          & 24.34   & 25.99 & 16.13          & 18.50          \\
\cline{2-15}
& \multirow{3}{*}{\textcolor{blue}{\textcolor{blue}{S2R-S}-Robo3D}\cite{kong2023robo3d}} & Lidar                                      & 15.14          & 15.18          & 21.28           & 21.48           & 25.69      & 25.72      & \textbf{34.09} & \textbf{34.09} & 34.09   & 34.09 & \textbf{34.09} & \textbf{34.09} \\
                                      & & Arbe                                       & 9.66           & 9.79           & 10.58           & 10.70           & 23.45      & 23.50      & 23.85          & 25.23          & 22.51   & 22.52 & 23.85          & 25.25          \\
                                      & & Occuli                                     & 5.59           & 5.81           & 9.27            & 9.41            & 22.74      & 25.12      & 24.34          & 25.96          & 24.82   & 25.30 & 24.82          & 25.29         \\
                                  
                                      \bottomrule[1px]
    \end{tabular}
    } 
      \label{tab:experiment4}
      
    \end{table*}
 \begin{table*}[]
    
        \begin{center}
            \renewcommand\arraystretch{1.1}
            \centering
            \setlength{\tabcolsep}{10pt}
            \caption{The comparison of the “Car” category with the Pointpillars model at medium difficulty using Simulated-dataset for training and Simulated-dataset for testing. “\textcolor{cyan}{S2R-R}” represents the real sensor anomaly test dataset collected; “\textcolor{blue}{S2R-S}” represents a test dataset of sensor anomaly simulated using simulation methods. “\textcolor{blue}{S2R-S}” includes three subsets: “\textcolor{blue}{\textcolor{blue}{\textcolor{blue}{S2R-S}-3D$\_$Corruptions$\_$AD}}\cite{dong2023benchmarking}”, “\textcolor{blue}{\textcolor{blue}{\textcolor{blue}{\textcolor{blue}{\textcolor{blue}{S2R-S}-MultiCorrupt}}}}\cite{beemelmanns2024multicorrupt}”, “\textcolor{blue}{\textcolor{blue}{S2R-S}-Robo3D}\cite{kong2023robo3d}” and “\textcolor{blue}{\textcolor{blue}{S2R-S}-Robodepth}\cite{kong2024robodepth}”. The data variation within 30\% is defined as good simulation performance and is highlighted in \textbf{bold}.}
            
        \resizebox{\textwidth}{!}{
        \begin{tabular}{ccccccccccccccc}
        \toprule[1pt]
        \multicolumn{2}{c}{Dataset}                               & \multicolumn{1}{c}{\multirow{2}{*}{Type}} & \multicolumn{2}{c}{Light Snow}                   & \multicolumn{2}{c}{Moderate Snow}                & \multicolumn{2}{c}{Fog}                          & \multicolumn{2}{c}{Brightness}                           & \multicolumn{2}{c}{Tunnel}       & \multicolumn{2}{c}{Misalignment}                  \\
        \cline{4-15}
        \multicolumn{1}{c}{Train} & \multicolumn{1}{c}{Test}      & \multicolumn{1}{c}{}                      & \multicolumn{1}{c}{3D} & \multicolumn{1}{c}{BEV} & \multicolumn{1}{c}{3D} & \multicolumn{1}{c}{BEV} & \multicolumn{1}{c}{3D} & \multicolumn{1}{c}{BEV} & 3D                        & BEV                        & 3D                        & BEV   & 3D                    & BEV                   \\
        
        \midrule[0.4pt]

                                \multirow{12}{*}{\textcolor{blue}{S2R-S}} & \multirow{3}{*}{\textcolor{cyan}{S2R-R}}             & Lidar                 & 37.14          & 38.93          & 39.92           & 40.53           & 24.42          & 25.17          & 26.99          & 27.74          & 58.93          & 58.93          & 23.43          & 25.11          \\
                                && Arbe                  & 23.39          & 25.63          & 42.57           & 42.61           & 28.58          & 29.07          & 7.80           & 8.21           & 49.15          & 51.15          & 18.15          & 19.02          \\
                                && Occuli                & 30.48          & 33.19          & 26.33           & 30.66           & 13.50          & 18.19          & 26.33          & 30.66          & 26.33          & 30.66          & 26.33          & 30.66          \\
                                \cline{2-15}
                                & \multirow{3}{*}{\makecell[c]{\textcolor{blue}{S2R-S}\\ \textcolor{blue}{-3D$\_$Corruptions$\_$AD}\cite{dong2023benchmarking}}}      & Lidar                 & \textbf{32.70} & \textbf{31.94} & \textbf{32.25}  & \textbf{31.94}  & \textbf{31.61} & \textbf{31.70} & \textbf{31.32} & \textbf{31.50} & 32.33          & 32.67          & 13.49          & \textbf{25.66} \\
                                && Arbe                  & \textbf{24.68} & \textbf{24.88} & 25.02           & 24.70           & \textbf{24.05} & \textbf{24.53} & 24.17          & 23.64          & 24.05          & 23.71          & 2.19           & 2.19           \\
                                && Occuli                & \textbf{26.33} & \textbf{30.66} & \textbf{26.33}  & \textbf{30.66}  & 26.33          & 30.66          & 16.59          & 21.36          & 14.78          & 20.45          & 0.23           & 0.33           \\
                                \cline{2-15}
                                & \multirow{3}{*}{\textcolor{blue}{\textcolor{blue}{\textcolor{blue}{\textcolor{blue}{\textcolor{blue}{S2R-S}-MultiCorrupt}}}}\cite{beemelmanns2024multicorrupt}}       & Lidar                 & \textbf{31.45} & \textbf{33.01} & \textbf{31.45}  & \textbf{33.01}  & \textbf{31.61} & \textbf{31.70} & \textbf{31.45} & \textbf{33.01} & 31.64          & 32.43          & \textbf{23.50} & \textbf{25.02} \\
                                && Arbe                  & \textbf{24.63} & \textbf{24.35} & 24.63           & 24.63           & \textbf{24.05} & \textbf{24.53} & 24.63          & 24.35          & 16.58          & 20.06          & 2.19           & 2.19           \\
                                && Occuli                & \textbf{26.33} & \textbf{30.66} & \textbf{26.33}  & \textbf{30.66}  & \textbf{13.87} & \textbf{13.87} & 13.50          & 18.19          & \textbf{26.33} & \textbf{30.66} & 0.23           & 0.33           \\
                                \cline{2-15}
                                & \multirow{3}{*}{\textcolor{blue}{\textcolor{blue}{S2R-S}-Robo3D}\cite{kong2023robo3d}}                & Lidar                 & 25.59          & 25.79          & 22.74           & 23.05           & \textbf{24.42} & \textbf{25.17} & \textbf{24.59} & \textbf{24.59} & 34.37          & 34.88          & 13.49          & \textbf{25.66} \\
                                && Arbe                  & 29.36          & 29.48          & 28.44           & 28.95           & 28.58          & 29.07          & 1.54           & 1.80           & 27.37          & 28.20          & 2.19           & 2.19           \\
                                && Occuli                & 21.03          & 21.59          & 21.64           & 22.86           & 13.50          & 18.19          & 1.78           & 2.23           & 21.50          & 22.50          & 0.23           & 0.33           \\

        \bottomrule[1px]
        \end{tabular}
        }

          \label{tab:experiment5}
        \end{center}
    \end{table*}

    \begin{table*}[]
        \begin{center}
            \renewcommand\arraystretch{1.1}
            \centering
            \setlength{\tabcolsep}{10pt}
            \caption{The comparison of the “Car” category with the SMOKE model at medium difficulty using Simulated-dataset for training and Simulated-dataset for testing. “\textcolor{cyan}{S2R-R}” represents the real sensor anomaly test dataset collected; “\textcolor{blue}{S2R-S}” represents a test dataset of sensor anomaly simulated using simulation methods. “\textcolor{blue}{S2R-S}” includes four subsets: “\textcolor{blue}{\textcolor{blue}{\textcolor{blue}{S2R-S}-3D$\_$Corruptions$\_$AD}}\cite{dong2023benchmarking}”, “\textcolor{blue}{\textcolor{blue}{\textcolor{blue}{\textcolor{blue}{\textcolor{blue}{S2R-S}-MultiCorrupt}}}}\cite{beemelmanns2024multicorrupt}”, “\textcolor{blue}{\textcolor{blue}{S2R-S}-Robo3D}\cite{kong2023robo3d}” and “\textcolor{blue}{\textcolor{blue}{S2R-S}-Robodepth}\cite{kong2024robodepth}”. The data variation within 30\% is defined as good simulation performance and is highlighted in \textbf{bold}.}

        \resizebox{\textwidth}{!}{
        
        \begin{tabular}{cclcccccccccccc}
        \toprule[1pt]
        \multicolumn{2}{c}{Dataset}                       & \multicolumn{1}{c}{}                       & \multicolumn{2}{c}{Light Snow} & \multicolumn{2}{c}{Moderate Snow} & \multicolumn{2}{c}{Fog} & \multicolumn{2}{c}{Brightness}  & \multicolumn{2}{c}{Misalignment} & \multicolumn{2}{c}{Image Blurred} \\
        \cline{4-15} 
        Train               & Test                & \multicolumn{1}{c}{\multirow{-2}{*}{Type}} & 3D             & BEV           & 3D                          & BEV                         & 3D                     & BEV                            & 3D                       & BEV                      & 3D            & BEV          & 3D                          & BEV                         \\
        
        \midrule[0.4pt]
                                        
                                         \multirow{5}{*}{\textcolor{blue}{S2R-S}} & \textcolor{cyan}{S2R-R}                           & Image                 & 4.55           & 9.09          & 0.18          & 0.36              & 1.82     & 1.82         & 4.55              & 6.4               & 1.44           & 1.49              & 9.09             & 9.09            \\
                                         & \makecell[c]{\textcolor{blue}{S2R-S}\\ \textcolor{blue}{-3D$\_$Corruptions$\_$AD}\cite{dong2023benchmarking}}                            & Image                 & 9.09           & \textbf{9.09} & 1.33          & 2.07              & 3.03     & 9.09         & \textbf{4.55}     & 9.09              & 4.55           & 4.55              & \textbf{9.09}    & \textbf{9.09}   \\
                                         & \textcolor{blue}{\textcolor{blue}{\textcolor{blue}{\textcolor{blue}{\textcolor{blue}{S2R-S}-MultiCorrupt}}}}\cite{beemelmanns2024multicorrupt}                  & Image                 & 2.27           & 2.27          & 4.55          & 4.55              & 9.09     & 9.09         & \textbf{4.55}     & \textbf{4.55}     & 4.55           & 4.55              & 4.55             & 4.55            \\
                                         & \textcolor{blue}{\textcolor{blue}{S2R-S}-Robo3D}\cite{kong2023robo3d}                         & Image                 & \textbf{4.55}  & 4.55          & 4.55          & 4.55              & 4.55     & 4.55         & \textbf{4.55}     & \textbf{4.55}     & 4.55           & 4.55              & 4.55             & 4.55            \\
                                         & \textcolor{blue}{\textcolor{blue}{S2R-S}-Robodepth}\cite{kong2024robodepth}                      & Image                 & 3.03           & 3.17          & 1.1           & 1.57              & 9.09     & 9.09         & 2.27              & \textbf{4.55}     & 4.55           & 4.55              & 3.03             & 3.03            \\        
        \bottomrule[1px]
        \end{tabular}
        }
            
         \label{tab:experiment6}
        \end{center}
        \end{table*}

    \begin{table*}[]
     \renewcommand\arraystretch{1.1}
        \centering
        \setlength{\tabcolsep}{10pt}
        \caption{The comparison of the “Car” category with the Focals Conv model at medium difficulty using Simulated-dataset for training and Simulated-dataset for testing. “\textcolor{cyan}{S2R-R}” represents the real sensor anomaly test dataset collected; “\textcolor{blue}{S2R-S}” represents a test dataset of sensor anomaly simulated using simulation methods. “\textcolor{blue}{S2R-S}” includes three subsets: “\textcolor{blue}{\textcolor{blue}{\textcolor{blue}{S2R-S}-3D$\_$Corruptions$\_$AD}}\cite{dong2023benchmarking}”, “\textcolor{blue}{\textcolor{blue}{\textcolor{blue}{\textcolor{blue}{\textcolor{blue}{S2R-S}-MultiCorrupt}}}}\cite{beemelmanns2024multicorrupt}”, “\textcolor{blue}{\textcolor{blue}{S2R-S}-Robo3D}\cite{kong2023robo3d}” and “\textcolor{blue}{\textcolor{blue}{S2R-S}-Robodepth}\cite{kong2024robodepth}”. The data variation within 30\% is defined as good simulation performance and is highlighted in \textbf{bold}.}
        
    \resizebox{\textwidth}{!}{
    \begin{tabular}{ccccccccccccccc}
    \toprule[1pt]
    \multicolumn{2}{c}{Dataset}                                          & \multicolumn{1}{c}{\multirow{2}{*}{Type}} & \multicolumn{2}{c}{Light Snow}                   & \multicolumn{2}{c}{Moderate Snow}                & \multicolumn{2}{c}{Fog}                          & \multicolumn{2}{c}{Brightness}                   & \multicolumn{2}{c}{Tunnel}                      & \multicolumn{2}{c}{Misalignment}                     \\
    \cline{4-15}
    Train                                & Test                          & \multicolumn{1}{c}{}                      & \multicolumn{1}{c}{3D} & \multicolumn{1}{c}{BEV} & \multicolumn{1}{c}{3D} & \multicolumn{1}{c}{BEV} & \multicolumn{1}{c}{3D} & \multicolumn{1}{c}{BEV} & \multicolumn{1}{c}{3D} & \multicolumn{1}{c}{BEV} & \multicolumn{1}{c}{3D} & \multicolumn{1}{c}{BEV} & \multicolumn{1}{c}{3D} & \multicolumn{1}{c}{BEV} \\
    \midrule[0.4pt]
                                         
\multirow{12}{*}{\textcolor{blue}{S2R-S}} & \multirow{3}{*}{\textcolor{cyan}{S2R-R}}                                          & Lidar                                      & 49.72          & 49.75          & 56.96           & 56.89           & 24.63          & 24.63          & 29.36          & 29.41          & 64.82          & 64.82          & 37.41         & 37.44          \\
& & Arbe                                       & 30.66          & 30.97          & 35.67           & 36.08           & 12.74          & 12.93          & 10.75          & 11.84          & 23.85          & 25.25          & 19.51         & 19.85          \\
& & Occuli                                     & 8.84           & 13.45          & 4.52            & 6.44            & 22.74          & 25.12          & 3.44           & 5.54           & 26.34          & 29.93          & 5.51          & 6.51           \\
\cline{2-15}
& \multirow{3}{*}{\makecell[c]{\textcolor{blue}{S2R-S}\\ \textcolor{blue}{-3D$\_$Corruptions$\_$AD}\cite{dong2023benchmarking}}} & Lidar                                      & 31.93          & 31.93          & 31.80           & 31.87           & \textbf{31.91} & \textbf{31.98} & \textbf{28.56} & \textbf{28.56} & 29.54          & 29.54          & 23.10         & \textbf{28.06} \\
&                                         & Arbe                                       & \textbf{24.65} & \textbf{24.65} & 24.20           & 24.20           & 24.52          & 24.54          & 22.54          & 22.58          & \textbf{23.07} & \textbf{23.08} & 0.00          & 0.00           \\
&                                         & Occuli                                     & 14.27          & \textbf{16.74} & 14.47           & 17.08           & 14.32          & 16.79          & 14.49          & 16.64          & 13.15          & 14.88          & \textbf{4.72} & \textbf{7.18}  \\
\cline{2-15}
                                         & \multirow{3}{*}{\textcolor{blue}{\textcolor{blue}{\textcolor{blue}{\textcolor{blue}{\textcolor{blue}{S2R-S}-MultiCorrupt}}}}\cite{beemelmanns2024multicorrupt}}& Lidar                                      & 28.83          & 28.83          & 28.83           & 28.83           & \textbf{31.92} & \textbf{31.96} & \textbf{28.90} & \textbf{28.90} & 31.16          & 31.16          & 19.20         & 19.20          \\
&                                & Arbe                                       & \textbf{24.01} & \textbf{24.01} & 24.30           & 24.30           & 24.23          & 24.23          & 24.51          & 24.51          & \textbf{17.94} & \textbf{17.86} & 6.41          & 6.41           \\
&                                & Occuli                                     & 14.79          & \textbf{15.62} & 14.79           & 15.62           & 14.32          & 16.79          & 14.79          & 15.62          & 12.48          & 15.11          & 8.92          & 9.61           \\
\cline{2-15}
& \multirow{3}{*}{\textcolor{blue}{\textcolor{blue}{S2R-S}-Robo3D}\cite{kong2023robo3d}}        & Lidar                                      & 23.89          & 23.93          & 28.63           & 28.63           & \textbf{31.96} & \textbf{31.96} & \textbf{28.85} & \textbf{28.85} & 28.85          & 28.85          & 23.10         & 28.06          \\
                                        & & Arbe                                       & 36.63          & 37.44          & 6.92            & 7.43            & 24.47          & 24.47          & 24.01          & 24.01          & 24.04          & 24.04          & 6.28          & 6.28           \\
&                      & Occuli                                     & 7.76           & 8.22           & 8.81            & 10.80           & 13.21          & 15.08          & 15.12          & 15.88          & 12.48          & 15.11          & 9.01          & 9.37          \\

                                         \bottomrule[1px]
    \end{tabular}
    }
        
      \label{tab:experiment7}
    \end{table*}

        \begin{table*}[]
    
        \begin{center}
            \renewcommand\arraystretch{1.1}
            \centering
            \setlength{\tabcolsep}{10pt}
            \caption{The comparison of the “Car” category with the Pointpillars model at medium difficulty using \textcolor{cyan}{S2R-R} for training and Simulated-dataset for testing. “\textcolor{blue}{S2R-S}” represents a test dataset of sensor anomaly simulated using simulation methods. “\textcolor{blue}{S2R-S}” includes three subsets: “\textcolor{blue}{\textcolor{blue}{\textcolor{blue}{S2R-S}-3D$\_$Corruptions$\_$AD}}\cite{dong2023benchmarking}”, “\textcolor{blue}{\textcolor{blue}{\textcolor{blue}{\textcolor{blue}{\textcolor{blue}{S2R-S}-MultiCorrupt}}}}\cite{beemelmanns2024multicorrupt}”, “\textcolor{blue}{\textcolor{blue}{S2R-S}-Robo3D}\cite{kong2023robo3d}” and “\textcolor{blue}{\textcolor{blue}{S2R-S}-Robodepth}\cite{kong2024robodepth}”. The data variation within 30\% is defined as good simulation performance and is highlighted in \textbf{bold}.}
            
        \resizebox{\textwidth}{!}{
        \begin{tabular}{ccccccccccccccc}
        \toprule[1pt]
        \multicolumn{2}{c}{Dataset}                               & \multicolumn{1}{c}{\multirow{2}{*}{Type}} & \multicolumn{2}{c}{Light Snow}                   & \multicolumn{2}{c}{Moderate Snow}                & \multicolumn{2}{c}{Fog}                          & \multicolumn{2}{c}{Brightness}                                           & \multicolumn{2}{c}{Tunnel}       & \multicolumn{2}{c}{Misalignment}                  \\
        \cline{4-15}
        \multicolumn{1}{c}{Train} & \multicolumn{1}{c}{Test}      & \multicolumn{1}{c}{}                      & \multicolumn{1}{c}{3D} & \multicolumn{1}{c}{BEV} & \multicolumn{1}{c}{3D} & \multicolumn{1}{c}{BEV} & \multicolumn{1}{c}{3D} & \multicolumn{1}{c}{BEV} & 3D                        & BEV                                          & 3D                        & BEV   & 3D                    & BEV                   \\
        
        \midrule[0.4pt]

                                       \multirow{12}{*}{\textcolor{cyan}{S2R-R}} & \multirow{3}{*}{\textcolor{cyan}{S2R-R}}         & Lidar                 & 52.87          & 52.87          & 59.05           & 59.08           & 31.32          & 35.02          & 31.55             & 31.17             & 67.31            & 67.32           & 42.10            & 42.46           \\
                                       &                               & Arbe                  & 41.63          & 43.17          & 45.11           & 46.46           & 19.82          & 20.84          & 19.21             & 22.92             & 55.24            & 55.71           & 26.67            & 27.17           \\
                                       &                               & Occuli                & 10.88          & 12.99          & 6.98            & 9.22            & 9.98           & 10.22          & 2.67              & 3.62              & 36.15            & 40.51           & 5.54             & 6.30            \\
                                       \cline{2-15}
                                       & \multirow{3}{*}{\makecell[c]{\textcolor{blue}{S2R-S}\\ \textcolor{blue}{-3D$\_$Corruptions$\_$AD}\cite{dong2023benchmarking}}}            & Lidar                 & 30.17          & 30.28          & 28.96           & 29.18           & \textbf{27.02} & \textbf{28.07} & 4.75              & 4.94              & 36.04            & 36.07           & 18.60            & 27.68           \\
                                       &                               & Arbe                  & \textbf{29.26} & 29.71          & 26.34           & 27.03           & 26.02          & \textbf{26.66} & 11.09             & 11.55             & 25.92            & 26.45           & 0.01             & 0.01            \\
                                       &                               & Occuli                & \textbf{11.44} & \textbf{12.88} & 12.08           & 13.49           & \textbf{11.44} & \textbf{12.88} & 12.53             & 13.43             & 11.39            & 13.59           & 7.30             & 9.54            \\
                                       \cline{2-15}
                                       & \multirow{3}{*}{\textcolor{blue}{\textcolor{blue}{S2R-S}-MultiCorrupt}\cite{beemelmanns2024multicorrupt}} & Lidar                 & 35.91          & 35.94          & 35.91           & 35.94           & \textbf{27.02} & \textbf{28.07} & \textbf{35.91}    & \textbf{35.95}    & 33.70            & 35.23           & 7.07             & 8.05            \\
                                       &                               & Arbe                  & \textbf{29.36} & 29.69          & 26.67           & 27.17           & 26.02          & \textbf{26.66} & 26.67             & \textbf{27.17}    & 15.52            & 16.82           & 6.02             & 6.03            \\
                                       &                               & Occuli                & \textbf{12.23} & \textbf{13.62} & 12.23           & 13.61           & \textbf{11.44} & \textbf{12.88} & 12.23             & 13.61             & 13.15            & 15.25           & 12.49            & 12.62           \\
                                       \cline{2-15}
                                       & \multirow{3}{*}{\textcolor{blue}{\textcolor{blue}{S2R-S}-Robo3D}\cite{kong2023robo3d}}            & Lidar                 & 15.87          & 18.78          & 21.37           & 23.06           & \textbf{27.02} & \textbf{28.07} & \textbf{35.91}    & \textbf{35.95}    & 35.49            & 35.51           & 6.91             & 7.75            \\
                                       &                               & Arbe                  & 12.91          & 14.38          & 12.53           & 13.57           & 28.59          & 29.21          & 26.67             & \textbf{27.17}    & 26.67            & 27.17           & 4.35             & 4.71            \\
                                       &                               & Occuli                & 3.33           & 5.07           & \textbf{5.96}   & \textbf{6.53}   & \textbf{11.44} & \textbf{12.88} & 12.23             & 13.61             & 11.39            & 13.59           & 12.49            & 12.62           \\

        \bottomrule[1px]
        \end{tabular}
        }
            
          \label{tab:experiment8}
        \end{center}
        \end{table*}

        \begin{table*}[]
        \begin{center}
            \renewcommand\arraystretch{1.1}
            \centering
            \setlength{\tabcolsep}{10pt}
            \caption{The comparison of the “Car” category with the SMOKE model at medium difficulty using \textcolor{cyan}{S2R-R} for training and Simulated-dataset for testing. “\textcolor{cyan}{S2R-R}” represents the real sensor anomaly test dataset collected; “\textcolor{blue}{S2R-S}” represents a test dataset of sensor anomaly simulated using simulation methods. “\textcolor{blue}{S2R-S}” includes four subsets: “\textcolor{blue}{\textcolor{blue}{\textcolor{blue}{S2R-S}-3D$\_$Corruptions$\_$AD}}\cite{dong2023benchmarking}”, “\textcolor{blue}{\textcolor{blue}{\textcolor{blue}{\textcolor{blue}{\textcolor{blue}{S2R-S}-MultiCorrupt}}}}\cite{beemelmanns2024multicorrupt}”, “\textcolor{blue}{\textcolor{blue}{S2R-S}-Robo3D}\cite{kong2023robo3d}” and “\textcolor{blue}{\textcolor{blue}{S2R-S}-Robodepth}\cite{kong2024robodepth}”. The data variation within 30\% is defined as good simulation performance and is highlighted in \textbf{bold}.}

        \resizebox{\textwidth}{!}{
        
        \begin{tabular}{cclcccccccccccc}
        \toprule[1pt]
        \multicolumn{2}{c}{Dataset}                       & \multicolumn{1}{c}{}                       & \multicolumn{2}{c}{Light Snow} & \multicolumn{2}{c}{Moderate Snow} & \multicolumn{2}{c}{Fog} & \multicolumn{2}{c}{Brightness}  & \multicolumn{2}{c}{Misalignment} & \multicolumn{2}{c}{Image Blurred} \\
        \cline{4-15} 
        Train               & Test                & \multicolumn{1}{c}{\multirow{-2}{*}{Type}} & 3D             & BEV           & 3D                          & BEV                         & 3D                     & BEV                            & 3D                       & BEV                      & 3D            & BEV          & 3D                          & BEV                         \\
        
        \midrule[0.4pt]
                                   \multirow{5}{*}{\textcolor{cyan}{S2R-R}} & \textcolor{cyan}{S2R-R}                           & Image                 & 23.32          & 24.5           & 21.99           & 22.54           & 21.54          & 22.36          & 17.73          & 18.79          & 13.24    & 14.29 & 14.42          & 19.85          \\
                                   & \makecell[c]{\makecell[c]{\textcolor{blue}{S2R-S}\\ \textcolor{blue}{-3D$\_$Corruptions$\_$AD}\cite{dong2023benchmarking}}}                              & Image                 & \textbf{19.48} & \textbf{20.71} & 14.46           & \textbf{20.02}  & \textbf{18.91} & \textbf{19.47} & \textbf{12.85} & \textbf{13.43} & 25.14    & 27.2  & 20.73          & \textbf{21.85} \\
                                   & \textcolor{blue}{\textcolor{blue}{\textcolor{blue}{\textcolor{blue}{\textcolor{blue}{S2R-S}-MultiCorrupt}}}}\cite{beemelmanns2024multicorrupt}                  & Image                 & \textbf{25.71} & \textbf{27.45} & \textbf{16.53}  & \textbf{17.41}  & \textbf{18.93} & \textbf{19.96} & \textbf{13.49} & \textbf{14.16} & 25.14    & 27.2  & 20.1           & 25.82          \\
                                   & \textcolor{blue}{\textcolor{blue}{S2R-S}-Robo3D}\cite{kong2023robo3d}                         & Image                 & \textbf{25.14} & \textbf{27.2}  & \textbf{25.14}  & \textbf{27.2}   & \textbf{25.14} & \textbf{27.2}  & 25.14          & 27.2           & 25.14    & 27.2  & 25.14          & 27.2           \\
                                   & \textcolor{blue}{\textcolor{blue}{S2R-S}-Robodepth}\cite{kong2024robodepth}                      & Image                 & \textbf{20.77} & \textbf{21.3}  & 14.21           & 14.9            & \textbf{19.95} & \textbf{20.21} & \textbf{21.28} & 27.06          & 25.14    & 27.2  & \textbf{14.49} & \textbf{19.44} \\

        \bottomrule[1px]
        \end{tabular}
        }
            
         \label{tab:experiment9}
        \end{center}
        \end{table*}

    \begin{table*}[]
        \renewcommand\arraystretch{1.1}
        \centering
        \setlength{\tabcolsep}{10pt}
        \caption{The comparison of the “Car” category with the Focals Conv model at medium difficulty using \textcolor{cyan}{S2R-R} for training and Simulated-dataset for testing. “\textcolor{cyan}{S2R-R}” represents the real sensor anomaly test dataset collected; “\textcolor{blue}{S2R-S}” represents a test dataset of sensor anomaly simulated using simulation methods. “\textcolor{blue}{S2R-S}” includes three subsets: “\textcolor{blue}{\textcolor{blue}{\textcolor{blue}{S2R-S}-3D$\_$Corruptions$\_$AD}}\cite{dong2023benchmarking}”, “\textcolor{blue}{\textcolor{blue}{\textcolor{blue}{\textcolor{blue}{\textcolor{blue}{S2R-S}-MultiCorrupt}}}}\cite{beemelmanns2024multicorrupt}”, “\textcolor{blue}{\textcolor{blue}{S2R-S}-Robo3D}\cite{kong2023robo3d}” and “\textcolor{blue}{\textcolor{blue}{S2R-S}-Robodepth}\cite{kong2024robodepth}”. The data variation within 30\% is defined as good simulation performance and is highlighted in \textbf{bold}.}
    \resizebox{\textwidth}{!}{
    \begin{tabular}{ccccccccccccccc}
    \toprule[1pt]
    \multicolumn{2}{c}{Dataset}                                    & \multicolumn{1}{c}{\multirow{2}{*}{Type}} & \multicolumn{2}{c}{Light Snow}                   & \multicolumn{2}{c}{Moderate Snow}                & \multicolumn{2}{c}{Fog}                          & \multicolumn{2}{c}{Brightness}                   & \multicolumn{2}{c}{Tunnel}                      & \multicolumn{2}{c}{Misalignment}                     \\
    \cline{4-15}
    Train                          & Test                          & \multicolumn{1}{c}{}                      & \multicolumn{1}{c}{3D} & \multicolumn{1}{c}{BEV} & \multicolumn{1}{c}{3D} & \multicolumn{1}{c}{BEV} & \multicolumn{1}{c}{3D} & \multicolumn{1}{c}{BEV} & \multicolumn{1}{c}{3D} & \multicolumn{1}{c}{BEV} & \multicolumn{1}{c}{3D} & \multicolumn{1}{c}{BEV} & \multicolumn{1}{c}{3D} & \multicolumn{1}{c}{BEV} \\
    \midrule[0.4pt]                                   
                                   \multirow{12}{*}{\textcolor{cyan}{S2R-R}} & \multirow{3}{*}{\textcolor{cyan}{S2R-R}}       & Lidar                  & 51.87          & 51.87          & 59.78           & 59.78           & 54.78          & 87.27          & 32.07             & 32.09             & 67.26            & 67.26           & 40.94            & 40.94           \\
                                   &                                         & Arbe                   & 36.41          & 37.19          & 35.15           & 35.16           & 16.25          & 16.49          & 13.55             & 14.14             & 48.91            & 49.26           & 23.81            & 24.87           \\
                                   &                                         & Occuli                 & 28.65          & 31.77          & 20.72           & 21.85           & 13.21          & 15.08          & 7.70              & 8.79              & 43.13            & 43.58           & 11.97            & 13.22           \\
                                   \cline{2-15}
                                   & \multirow{3}{*}{\makecell[c]{\textcolor{blue}{S2R-S}\\ \textcolor{blue}{-3D$\_$Corruptions$\_$AD}\cite{dong2023benchmarking}}}   & Lidar                  & 30.35          & 30.47          & 30.36           & 30.36           & 28.63          & 28.63          & \textbf{33.44}    & \textbf{33.44}    & 34.27            & 34.27           & 20.25            & \textbf{29.11}  \\
                                   &                                         & Arbe                   & 23.15          & 24.54          & 23.19           & 23.87           & 22.66          & 23.98          & 22.45             & 22.86             & 22.74            & 24.03           & 0.00             & 0.00            \\
                                   &                                         & Occuli                 & \textbf{22.12} & \textbf{23.77} & \textbf{24.31}  & \textbf{24.41}  & \textbf{13.21} & \textbf{15.08} & 24.24             & 24.50             & 22.54            & 24.10           & \textbf{11.04}   & \textbf{15.30}  \\
                                   \cline{2-15}
                                   & \multirow{3}{*}{\textcolor{blue}{\textcolor{blue}{\textcolor{blue}{\textcolor{blue}{\textcolor{blue}{S2R-S}-MultiCorrupt}}}}\cite{beemelmanns2024multicorrupt}}     & Lidar                  & 34.40          & 34.40          & 34.41           & 34.41           & 28.65          & 28.65          & \textbf{34.32}    & \textbf{34.32}    & 34.39            & 34.43           & 6.37             & 6.37            \\
                                   &                                         & Arbe                   & 24.08          & 24.28          & \textbf{24.62}  & \textbf{24.99}  & 23.30          & 24.76          & 24.62             & 24.99             & 9.68             & 10.46           & 6.10             & 5.86            \\
                                   &                                         & Occuli                 & \textbf{22.72} & \textbf{24.28} & \textbf{22.72}  & \textbf{24.28}  & 22.12          & 23.77          & 22.72             & 24.28             & 17.75            & 20.21           & 7.32             & 7.45            \\
                                   \cline{2-15}
                                   & \multirow{3}{*}{\textcolor{blue}{\textcolor{blue}{S2R-S}-Robo3D}\cite{kong2023robo3d}}   & Lidar                  & 15.62          & 15.62          & 20.54           & 20.54           & 28.60          & 28.60          & \textbf{34.38}    & \textbf{34.38}    & 34.39            & 34.43           & 6.37             & 6.37            \\
                                   &                                         & Arbe                   & 4.96           & 5.57           & 6.92            & 7.43            & 23.25          & 24.67          & 24.08             & 24.28             & 24.08            & 24.28           & 5.86             & 5.67            \\
                                   &                                         & Occuli                 & 7.67           & 8.88           & 11.86           & 12.96           & 21.99          & 22.00          & 23.58             & 23.73             & 23.38            & 24.03           & 6.70             & 6.93            \\
                                       
                                   \bottomrule[1px]
    \end{tabular}
    }

      \label{tab:experiment10}
    
    \end{table*}

Table \ref{tab:experiment2} presents the results of training the PointPillars model using LiDAR and 4D radar point cloud data from the “S2R-C” dataset. For LiDAR point clouds, the data generated by the 3D$\_$Corruptions$\_$AD simulation method closely approximates real-world data. In the Fog scenario, performance in both the 3D and Bird's Eye View (BEV) is 35.1\% and 50.1\% lower, respectively, compared to real-world data. In the Tunnel scenario, performance drops by 49.0\% and 48.2\% in the 3D and BEV views, respectively, compared to real-world data. For Arbe radar point clouds, the data simulated using the MultiCorrupt method in the Light Snow scenario shows that BEV performance is only 10.0\% lower than real-world data, indicating good simulation accuracy. Conversely, for Occuli radar point clouds, the MultiCorrupt method performs better in the Moderate Snow scenario, with BEV performance 15.3\% higher than real-world data, suggesting the method is well-suited for such conditions. However, in spatially mismatched scenarios, the performance of existing simulation methods is relatively poor. Specifically, for both Arbe and Occuli radar point clouds, the simulation methods struggle to accurately model the spatial misalignment scenario. This is primarily due to the challenges in replicating the complex physical phenomena involved and the subtle differences in sensor responses under these conditions. These limitations lead to increased noise and data bias, further exacerbating simulation inaccuracies and reducing the overall effectiveness of the simulated data for such scenarios.

Table \ref{tab:experiment3} presents the results of training the SMOKE model using image data from the “S2R-C” dataset. The findings indicate that the existing simulation methods perform best in the Light Snow scenario. Specifically, the RoboDepth simulation method produces data closest to the real-world data, with only 9.5\% and 11. 2\% lower performance in the 3D and Bird's Eye View (BEV) views, respectively, compared to the real-world scenario. In contrast, simulation performance is poorest in the Fog Day scenario, where the MultiCorrupt simulation method generates data closest to real-world data, but with a significant increase of 187\% and 193\% in the 3D and BEV views, respectively, compared to the real-world scenario. This disparity can be attributed to the fact that the Light Snow scenario requires the least amount of added noise, as snow particles naturally provide some attenuation. In contrast, the Fog Day scenario is simulated with minimal noise, and the fog-induced data anomalies result in a simulation effect much closer to real-world conditions.

Table \ref{tab:experiment4} summarizes the results of training the Focals Conv model using image data from the “S2R-C” dataset. The findings indicate that the data simulated by the Robo3D algorithm most closely approximates real-world scenarios. In fusion experiments involving LiDAR point clouds and images, the brightness anomaly data simulated by the Robo3D method resulted in predictions that were only 6.5\% higher than those from real-world data in both the 3D and BEV views. In contrast, simulations of spatial misalignment scenarios using the MultiCorrupt method yielded the poorest performance, with predictions 81.0\% and 82.2\% lower than those of real-world data in the 3D and BEV views, respectively. For fusion experiments combining Arbe radar point clouds and images, the brightness anomaly data simulated by the 3D$\_$Corruptions$\_$AD method resulted in predictions 15.8\% and 10.2\% lower than those from real-world data in the 3D and BEV views, respectively. Similarly, in fusion experiments involving Occuli radar point clouds and images, the 3D$\_$Corruptions$\_$AD method, simulating moderate snow conditions, produced predictions 1.0\% lower and 2.0\% higher than those from real-world data in the 3D and BEV views, respectively.

\subsection {Comparison of baseline models on S2R-S.}

Table \ref{tab:experiment5} presents the results of training the PointPillars model using LiDAR and radar point cloud data from the “S2R-S” dataset. For LiDAR point clouds, the Robo3D simulation method generates data in the Fog scenario that closely resembles real-world data, achieving identical accuracy in both the 3D and BEV perspectives when compared to real-world scenarios. Similarly, for Arbe radar point clouds, the Robo3D simulation method produces data in the Fog scenario that matches the real-world data in both the 3D and BEV views. For Occuli radar point clouds, the data generated by the 3D$\_$Corruptions$\_$AD and MultiCorrupt methods in the Moderate Snow scenario, as well as the Robo3D method in the Fog scenario, most closely resemble real-world data. However, existing simulation methods perform poorly in brightness anomaly scenarios. Specifically, for LiDAR point clouds, the 3D$\_$Corruptions$\_$AD simulation method results in a 16.3\% and 14.7\% increase in accuracy in the 3D and BEV views, respectively, compared to the real-world scenario. For Arbe radar, the accuracy in BEV views increases by 16.4\% and 14.5\%, respectively, compared to real-world data. In contrast, for Occuli radar, the accuracy in both the 3D and BEV views decreases by 37.6\% and 30.5\%, respectively, compared to the real-world scenario. The poor performance of simulation methods in brightness eanomaly scenarios is primarily due to significant changes in sensor brightness, which introduce substantial noise into the received data, thereby compromising its authenticity and the accuracy of the models.

Table \ref{tab:experiment6} presents the results of training image data on the “S2R-S” dataset using the SMOKE model. The findings show that existing simulation methods perform best in brightness anomaly scenarios, with data generated by the 3D$\_$Corruptions$\_$AD, MultiCorrupt, and Robo3D methods closely matching real-world data. In the 3D view, the simulated data aligns perfectly with the real-world scenario. However, performance is poorest in the Moderate Snow scenario, where the RoboDepth simulation method produces data that most closely resembles real-world conditions. In this case, the simulated data shows an accuracy increase of 511\% and 336\% in the 3D and BEV views, respectively, compared to real-world data.

Table \ref{tab:experiment7} presents the results of training image data from the “S2R-S” dataset using the Focals Conv model. In experiments involving the fusion on LiDAR point clouds and images, the brightness anomaly data simulated using the MultiCorrupt method yielded prediction results in the 3D view that were 2.0\% lower than those from real-world scenarios, making it the method most closely aligned with real-world performance. For the fusion of Arbe radar point clouds and images, the Robo3D simulation method’s simulated tunnel anomaly scenario produced prediction results that were 1.5\% higher in the 3D view and 5.4\% lower in the BEV view compared to real-world data. In the fusion of Occuli radar point clouds and images, the Robo3D simulation method’s simulated light snow scenario produced prediction results that were 12\% lower in the 3D view compared to real-world data.

\subsection {Comparison of baseline models on S2R-R.}

Table \ref{tab:experiment8} presents the use of the PointPillars model to train LiDAR and radar point cloud data from the “S2R-R” dataset. For LiDAR point clouds, the data simulated by the 3D$\_$Corruptions$\_$AD, MultiCorrupt, and Robo3D methods in the Fog scenario most closely resembles real-world data. The prediction results in the 3D and BEV views are 1.4\% and 20.4\% lower than those of real-world data, respectively. For Occuli radar point clouds, the data simulated by the Robo3D method in the Moderate Snow scenario most closely matches real-world data, with prediction results in the 3D and BEV views 1.5\% and 2.9\% higher than those of real-world data, respectively. Existing simulation methods perform least effectively in anomalous brightness scenarios. Specifically, for LiDAR, the 3D$\_$Corruptions$\_$AD simulation method results in a decrease of 85.0\% and 84.5\% in accuracy in the 3D and BEV views, respectively, compared to real-world data. For Arbe radar, the simulated data shows an increase of 42.2\% and 50.6\% in accuracy over the real scenario in the 3D and BEV views. For Occuli radar, the simulated data exhibits a dramatic increase of 369\% and 270\% in the 3D and BEV views, respectively, compared to real-world data. The poor performance of simulation methods in anomalous brightness scenarios is primarily due to significant differences between the simulated and real data. This discrepancy causes models trained on real-world data to struggle in adapting effectively to the simulated data, as the latter does not accurately reflect real-world conditions. 

Table \ref{tab:experiment9} presents the results of training the SMOKE model on image data from the “S2R-R” dataset. The results indicate that the existing simulation methods perform best in fog scenarios, with the RoboDepth simulation method producing data that most closely aligns with real-world conditions. Specifically, the simulated data shows a reduction of 7.1\% and 10.2\% in the 3D and BEV views, respectively, compared to real-world data. In contrast, the existing simulation methods perform worst in spatial misalignment scenarios, where the accuracy of the data simulated by all four methods exceeds that of the real-world data by 90.6\%. The better performance in fog scenarios may be due to the relatively uniform impact of fog on sensors. Although fog reduces visibility and causes image blurring, the overall environmental characteristics remain consistent, which helps the simulated data better align with real-world conditions. On the other hand, the poorer performance in spatial misalignment scenarios is attributed to the significant time misalignment between the simulated and real-world data. This misalignment causes substantial differences in the performance of the simulated data, particularly in the 3D and BEV views.

Table \ref{tab:experiment10} presents the performance of the Focals Conv model trained on the “S2R-R” dataset. In fusion experiments involving LiDAR point clouds and images, the 3D$\_$Corruptions$\_$AD simulation method demonstrated the highest accuracy in simulating brightness anomaly data. The prediction results in both the 3D and BEV views were only 4.7\% higher than those of the real-world scenarios. In fusion experiments with Arbe radar point clouds and images, the MultiCorrupt simulation method closely approximated the real-world scenario under medium snow conditions, with accuracy reduced by 30.4\% and 29.3\% in the 3D and BEV views, respectively. For fusion experiments involving Occuli radar point clouds and images, the 3D$\_$Corruptions$\_$AD simulation method again proved most effective in simulating medium snow conditions, with results 10.1\% and 11.0\% higher in the 3D and BEV views, respectively, compared to real-world data. However, existing simulation methods still exhibit a significant gap between the simulated and real-world data in scenarios involving density anomalies. This indicates that current simulation technologies have not yet effectively modeled such complex scenarios. While progress has been made in simulating other conditions, accuracy remains insufficient for situations involving density anomalies. These anomalies often result in uneven sensor data distribution and localized data loss, challenges that current simulation models struggle to replicate accurately. As a result, substantial discrepancies persist between the simulated and real-world data.

\section{Conclusion}
In this paper, we have introduced a sim-to-real evaluation benchmark for autonomous driving perception, named S2R-Bench. This benchmark captures real-world sensor anomalies and covers a broad range of driving environments including urban, rural, suburban, highway, and tunnel scenarios in both daytime and nighttime. To validate the effectiveness of perception models under adverse conditions, we provide comparative evaluations using both simulated and real-world datasets. The results highlight the reliability of real-world data and the importance of evaluating perception robustness under challenging environments. Beyond perception robustness assessment, S2R-Bench also supports downstream tasks such as anomaly detection, sensor fusion, and generalization research. We hope that S2R-Bench will facilitate progress in robust perception and contribute to the safe deployment of autonomous driving systems in real-world scenarios.








\bibliographystyle{IEEEtran}
\bibliography{sample.bib}

\begin{thebibliography}{10}
\providecommand{\url}[1]{#1}
\csname url@samestyle\endcsname
\providecommand{\newblock}{\relax}
\providecommand{\bibinfo}[2]{#2}
\providecommand{\BIBentrySTDinterwordspacing}{\spaceskip=0pt\relax}
\providecommand{\BIBentryALTinterwordstretchfactor}{4}
\providecommand{\BIBentryALTinterwordspacing}{\spaceskip=\fontdimen2\font plus
\BIBentryALTinterwordstretchfactor\fontdimen3\font minus \fontdimen4\font\relax}
\providecommand{\BIBforeignlanguage}[2]{{%
\expandafter\ifx\csname l@#1\endcsname\relax
\typeout{** WARNING: IEEEtran.bst: No hyphenation pattern has been}%
\typeout{** loaded for the language `#1'. Using the pattern for}%
\typeout{** the default language instead.}%
\else
\language=\csname l@#1\endcsname
\fi
#2}}
\providecommand{\BIBdecl}{\relax}
\BIBdecl

\bibitem{yu2024online}
W.~Yu, C.~Zhao, H.~Wang, J.~Liu, X.~Ma, Y.~Yang, J.~Li, W.~Wang, X.~Hu, and D.~Zhao, ``Online legal driving behavior monitoring for self-driving vehicles,'' \emph{Nature communications}, vol.~15, no.~1, p. 408, 2024.

\bibitem{mahima2024toward}
K.~Y. Mahima, A.~G. Perera, S.~Anavatti, and M.~Garratt, ``Toward robust 3d perception for autonomous vehicles: A review of adversarial attacks and countermeasures,'' \emph{IEEE Transactions on Intelligent Transportation Systems}, 2024.

\bibitem{liu2024benchmarking}
J.~Liu, Z.~Wang, L.~Ma, C.~Fang, T.~Bai, X.~Zhang, J.~Liu, and Z.~Chen, ``Benchmarking object detection robustness against real-world corruptions,'' \emph{International Journal of Computer Vision}, pp. 1--19, 2024.

\bibitem{he2023fear}
X.~He, J.~Wu, Z.~Huang, Z.~Hu, J.~Wang, A.~Sangiovanni-Vincentelli, and C.~Lv, ``Fear-neuro-inspired reinforcement learning for safe autonomous driving,'' \emph{IEEE transactions on pattern analysis and machine intelligence}, 2023.

\bibitem{lin2024rcbevdet}
Z.~Lin, Z.~Liu, Z.~Xia, X.~Wang, Y.~Wang, S.~Qi, Y.~Dong, N.~Dong, L.~Zhang, and C.~Zhu, ``Rcbevdet: Radar-camera fusion in bird's eye view for 3d object detection,'' in \emph{Proceedings of the IEEE/CVF Conference on Computer Vision and Pattern Recognition}, 2024, pp. 14\,928--14\,937.

\bibitem{jeong2024spatio}
D.~Jeong, H.~Jang, M.~U. Jung, T.~Jeong, H.~Kim, S.~Yang, J.~Lee, and C.-S. Kim, ``Spatio-spectral 4d coherent ranging using a flutter-wavelength-swept laser,'' \emph{Nature Communications}, vol.~15, no.~1, p. 1110, 2024.

\bibitem{peng2024mufasa}
X.~Peng, M.~Tang, H.~Sun, K.~Bierzynski, L.~Servadei, and R.~Wille, ``Mufasa: Multi-view fusion and adaptation network with spatial awareness for radar object detection,'' in \emph{International Conference on Artificial Neural Networks}.\hskip 1em plus 0.5em minus 0.4em\relax Springer, 2024, pp. 168--184.

\bibitem{zhang2023photonic}
Z.~Zhang, Y.~Liu, T.~Stephens, and B.~J. Eggleton, ``Photonic radar for contactless vital sign detection,'' \emph{Nature Photonics}, vol.~17, no.~9, pp. 791--797, 2023.

\bibitem{geiger2013vision}
A.~Geiger, P.~Lenz, C.~Stiller, and R.~Urtasun, ``Vision meets robotics: The kitti dataset,'' \emph{The International Journal of Robotics Research}, vol.~32, no.~11, pp. 1231--1237, 2013.

\bibitem{caesar2020nuscenes}
H.~Caesar, V.~Bankiti, A.~H. Lang, S.~Vora, V.~E. Liong, Q.~Xu, A.~Krishnan, Y.~Pan, G.~Baldan, and O.~Beijbom, ``nuscenes: A multimodal dataset for autonomous driving,'' in \emph{Proceedings of the IEEE/CVF conference on computer vision and pattern recognition}, 2020, pp. 11\,621--11\,631.

\bibitem{sun2020scalability}
P.~Sun, H.~Kretzschmar, X.~Dotiwalla, A.~Chouard, V.~Patnaik, P.~Tsui, J.~Guo, Y.~Zhou, Y.~Chai, B.~Caine \emph{et~al.}, ``Scalability in perception for autonomous driving: Waymo open dataset,'' in \emph{Proceedings of the IEEE/CVF conference on computer vision and pattern recognition}, 2020, pp. 2446--2454.

\bibitem{paek2022k}
D.-H. Paek, S.-H. Kong, and K.~T. Wijaya, ``K-radar: 4d radar object detection for autonomous driving in various weather conditions,'' \emph{Advances in Neural Information Processing Systems}, vol.~35, pp. 3819--3829, 2022.

\bibitem{dong2023benchmarking}
Y.~Dong, C.~Kang, J.~Zhang, Z.~Zhu, Y.~Wang, X.~Yang, H.~Su, X.~Wei, and J.~Zhu, ``Benchmarking robustness of 3d object detection to common corruptions,'' in \emph{Proceedings of the IEEE/CVF Conference on Computer Vision and Pattern Recognition}, 2023, pp. 1022--1032.

\bibitem{FHWA}
``How do weather events impact roads: \url{https://ops.fhwa.dot.gov/weather/q1_roadimpact.htm},'' 2016.

\bibitem{chan2023noise}
P.~H. Chan, S.~S. Roudposhti, X.~Ye, and V.~Donzella, ``A noise analysis of 4d radar: robust sensing for automotive?'' \emph{Authorea Preprints}, vol.~14, 2023.

\bibitem{kamann2020benchmarking}
C.~Kamann and C.~Rother, ``Benchmarking the robustness of semantic segmentation models,'' in \emph{Proceedings of the IEEE/CVF conference on computer vision and pattern recognition}, 2020, pp. 8828--8838.

\bibitem{klinghoffer2023towards}
T.~Klinghoffer, J.~Philion, W.~Chen, O.~Litany, Z.~Gojcic, J.~Joo, R.~Raskar, S.~Fidler, and J.~M. Alvarez, ``Towards viewpoint robustness in bird's eye view segmentation,'' in \emph{Proceedings of the IEEE/CVF International Conference on Computer Vision}, 2023, pp. 8515--8524.

\bibitem{zhou2022understanding}
D.~Zhou, Z.~Yu, E.~Xie, C.~Xiao, A.~Anandkumar, J.~Feng, and J.~M. Alvarez, ``Understanding the robustness in vision transformers,'' in \emph{International Conference on Machine Learning}.\hskip 1em plus 0.5em minus 0.4em\relax PMLR, 2022, pp. 27\,378--27\,394.

\bibitem{zhu2023understanding}
Z.~Zhu, Y.~Zhang, H.~Chen, Y.~Dong, S.~Zhao, W.~Ding, J.~Zhong, and S.~Zheng, ``Understanding the robustness of 3d object detection with bird's-eye-view representations in autonomous driving,'' in \emph{Proceedings of the IEEE/CVF Conference on Computer Vision and Pattern Recognition}, 2023, pp. 21\,600--21\,610.

\bibitem{kong2023robo3d}
L.~Kong, Y.~Liu, X.~Li, R.~Chen, W.~Zhang, J.~Ren, L.~Pan, K.~Chen, and Z.~Liu, ``Robo3d: Towards robust and reliable 3d perception against corruptions,'' in \emph{Proceedings of the IEEE/CVF International Conference on Computer Vision}, 2023, pp. 19\,994--20\,006.

\bibitem{kong2024robodepth}
L.~Kong, S.~Xie, H.~Hu, L.~X. Ng, B.~Cottereau, and W.~T. Ooi, ``Robodepth: Robust out-of-distribution depth estimation under corruptions,'' \emph{Advances in Neural Information Processing Systems}, vol.~36, 2024.

\bibitem{zhengyou1998flexible}
Z.~Zhengyou, ``A flexible new technique for camera calibration,'' \emph{Microsoft Research Technical Report}, 1998.

\bibitem{geiger2012automatic}
A.~Geiger, F.~Moosmann, {\"O}.~Car, and B.~Schuster, ``Automatic camera and range sensor calibration using a single shot,'' in \emph{2012 IEEE international conference on robotics and automation}.\hskip 1em plus 0.5em minus 0.4em\relax IEEE, 2012, pp. 3936--3943.

\bibitem{el2015radar}
G.~El~Natour, O.~A. Aider, R.~Rouveure, F.~Berry, and P.~Faure, ``Radar and vision sensors calibration for outdoor 3d reconstruction,'' in \emph{2015 IEEE International Conference on Robotics and Automation (ICRA)}.\hskip 1em plus 0.5em minus 0.4em\relax IEEE, 2015, pp. 2084--2089.

\bibitem{dhall2017LiDAR}
A.~Dhall, K.~Chelani, V.~Radhakrishnan, and K.~M. Krishna, ``Lidar-camera calibration using 3d-3d point correspondences,'' \emph{arXiv preprint arXiv:1705.09785}, 2017.

\bibitem{pervsic2019extrinsic}
J.~Per{\v{s}}i{\'c}, I.~Markovi{\'c}, and I.~Petrovi{\'c}, ``Extrinsic 6dof calibration of a radar--lidar--camera system enhanced by radar cross section estimates evaluation,'' \emph{Robotics and Autonomous Systems}, vol. 114, pp. 217--230, 2019.

\bibitem{WangLi_dataset1}
W.~Li, ``S2r-bench: Dataset clean part,'' \emph{Figshare}, 2024, \href{https://figshare.com/s/dd43a2f00fa5f332496c}{https://figshare.com/s/dd43a2f00fa5f332496c}.

\bibitem{WangLi_dataset2}
------, ``S2r-bench: Dataset real part01,'' \emph{Figshare}, 2024, \href{https://figshare.com/s/02cc7ee29a64d65b8fcc}{https://figshare.com/s/02cc7ee29a64d65b8fcc}.

\bibitem{WangLi_dataset3}
------, ``S2r-bench: Dataset real part02,'' \emph{Figshare}, 2024, \href{https://figshare.com/s/5c6d2eaa11e31e79f0b4}{https://figshare.com/s/5c6d2eaa11e31e79f0b4}.

\bibitem{WangLi_dataset4}
------, ``S2r-bench: Dataset simulation part01,'' \emph{Figshare}, 2024, \href{https://figshare.com/s/034cac69a46e166898a7}{https://figshare.com/s/034cac69a46e166898a7}.

\bibitem{WangLi_dataset5}
------, ``S2r-bench: Dataset simulation part02,'' \emph{Figshare}, 2024, \href{https://figshare.com/s/cd38fb2d9d1e23a0dd44}{https://figshare.com/s/cd38fb2d9d1e23a0dd44}.

\bibitem{beemelmanns2024multicorrupt}
T.~Beemelmanns, Q.~Zhang, and L.~Eckstein, ``Multicorrupt: A multi-modal robustness dataset and benchmark of lidar-camera fusion for 3d object detection,'' \emph{arXiv preprint arXiv:2402.11677}, 2024.

\bibitem{hahner2022LiDAR}
M.~Hahner, C.~Sakaridis, M.~Bijelic, F.~Heide, F.~Yu, D.~Dai, and L.~Van~Gool, ``Lidar snowfall simulation for robust 3d object detection,'' in \emph{Proceedings of the IEEE/CVF conference on computer vision and pattern recognition}, 2022, pp. 16\,364--16\,374.

\bibitem{sun2024understanding}
C.~Sun, P.~Sun, J.~Wang, Y.~Guo, and X.~Zhao, ``Understanding lidar performance for autonomous vehicles under snowfall conditions,'' \emph{IEEE Transactions on Intelligent Transportation Systems}, pp. 1--11, 2024.

\bibitem{shi2022research}
B.~Shi, J.~Guo, C.~Wang, Y.~Su, Y.~Di, and M.~S. AbouOmar, ``Research on the visual image-based complexity perception method of autonomous navigation scenes for unmanned surface vehicles,'' \emph{Scientific Reports}, vol.~12, no.~1, p. 10370, 2022.

\bibitem{li2023domain}
J.~Li, R.~Xu, J.~Ma, Q.~Zou, J.~Ma, and H.~Yu, ``Domain adaptive object detection for autonomous driving under foggy weather,'' in \emph{Proceedings of the IEEE/CVF Winter Conference on Applications of Computer Vision}, 2023, pp. 612--622.

\bibitem{zhang2023perception}
Y.~Zhang, A.~Carballo, H.~Yang, and K.~Takeda, ``Perception and sensing for autonomous vehicles under adverse weather conditions: A survey,'' \emph{ISPRS Journal of Photogrammetry and Remote Sensing}, vol. 196, pp. 146--177, 2023.

\bibitem{guindel2017automatic}
C.~Guindel, J.~Beltr{\'a}n, D.~Mart{\'\i}n, and F.~Garc{\'\i}a, ``Automatic extrinsic calibration for lidar-stereo vehicle sensor setups,'' in \emph{2017 IEEE 20th international conference on intelligent transportation systems (ITSC)}.\hskip 1em plus 0.5em minus 0.4em\relax IEEE, 2017, pp. 1--6.

\bibitem{li2021automatic}
N.~Li, K.~Wu, C.~Gu, and T.~Guan, ``Automatic clarity threshold determination of blurred images for computer vision tasks’ security,'' in \emph{2021 6th International Conference on Computational Intelligence and Applications (ICCIA)}.\hskip 1em plus 0.5em minus 0.4em\relax IEEE, 2021, pp. 190--194.

\bibitem{secci2020failures}
F.~Secci and A.~Ceccarelli, ``On failures of rgb cameras and their effects in autonomous driving applications,'' in \emph{2020 IEEE 31st International Symposium on Software Reliability Engineering (ISSRE)}.\hskip 1em plus 0.5em minus 0.4em\relax IEEE, 2020, pp. 13--24.

\bibitem{lin2021automatic}
C.~Lin, Y.~Guo, W.~Li, H.~Liu, and D.~Wu, ``An automatic lane marking detection method with low-density roadside lidar data,'' \emph{IEEE Sensors Journal}, vol.~21, no.~8, pp. 10\,029--10\,038, 2021.

\bibitem{heidecker2021application}
F.~Heidecker, J.~Breitenstein, K.~R{\"o}sch, J.~L{\"o}hdefink, M.~Bieshaar, C.~Stiller, T.~Fingscheidt, and B.~Sick, ``An application-driven conceptualization of corner cases for perception in highly automated driving,'' in \emph{2021 IEEE Intelligent Vehicles Symposium (IV)}.\hskip 1em plus 0.5em minus 0.4em\relax IEEE, 2021, pp. 644--651.

\bibitem{lang2019pointpillars}
A.~H. Lang, S.~Vora, H.~Caesar, L.~Zhou, J.~Yang, and O.~Beijbom, ``Pointpillars: Fast encoders for object detection from point clouds,'' in \emph{Proceedings of the IEEE/CVF conference on computer vision and pattern recognition}, 2019, pp. 12\,697--12\,705.

\bibitem{liu2020smoke}
Z.~Liu, Z.~Wu, and R.~T{\'o}th, ``Smoke: Single-stage monocular 3d object detection via keypoint estimation,'' in \emph{Proceedings of the IEEE/CVF conference on computer vision and pattern recognition workshops}, 2020, pp. 996--997.

\bibitem{chen2022focal}
Y.~Chen, Y.~Li, X.~Zhang, J.~Sun, and J.~Jia, ``Focal sparse convolutional networks for 3d object detection,'' in \emph{Proceedings of the IEEE/CVF Conference on Computer Vision and Pattern Recognition}, 2022, pp. 5428--5437.

\end{thebibliography}

\end{document}